%% file: TRO2021.tex
\documentclass[journal]{IEEEtran}

\makeatletter
\def\endthebibliography{%
	\def\@noitemerr{\@latex@warning{Empty `thebibliography' environment}}%
	\endlist
}
\makeatother

\usepackage{cite}
\ifCLASSINFOpdf
  \usepackage[pdftex]{graphicx}
\else
\fi
\usepackage{amsmath}
\usepackage{amssymb}
\usepackage{wasysym}
\usepackage{multirow}
\usepackage{enumitem}   
\usepackage{color, colortbl}
\usepackage{xcolor}
\definecolor{Gray}{gray}{0.9}
\usepackage{hyperref}

\usepackage{amsthm}
\newtheorem{theorem}{Theorem}
\newtheorem{corollary}{Corollary}

\newtheorem{proposition}{Proposition}
\newtheorem{definition}{Definition}
\newtheorem{remark}{Remark}
\DeclareMathOperator*{\nX}{nX}
\DeclareMathOperator*{\diag}{diag}

\begin{document}

\title{Guarantees for Real Robotic Systems: \\ Unifying Formal Controller Synthesis and Reachset-Conformant Identification}

\author{Stefan~B.~Liu,
		Bastian~Sch\"urmann, and
        Matthias~Althoff
\thanks{All authors are with the Department of Informatics, Technical University of Munich, Garching,
85748, Germany. Email: [stefan.liu; bastian.schuermann; althoff]@tum.de.}
\thanks{Manuscript received April XX, XXXX; revised August XX, XXXX. This work was supported by the European Union’s Horizon 2020 Research and Innovation Program under Grant Agreement 101016007 (Project CONCERT).}}

\markboth{Transactions on Robotics,~Vol.~X, No.~X, August~XXXX}%
{Shell \MakeLowercase{\textit{et al.}}: Bare Demo of IEEEtran.cls for IEEE Journals}

\maketitle

\begin{abstract}
	Robots are used increasingly often in safety-critical scenarios, such as robotic surgery or human-robot interaction. To ensure stringent performance criteria, formal controller synthesis is a promising direction to guarantee that robots behave as desired. However, formally ensured properties only transfer to the real robot when the model is appropriate. We address this problem by combining the identification of a reachset-conformant model with controller synthesis. Since the reachset-conformant model contains all the measured behaviors of the real robot, the safety properties of the model transfer to the real robot. The transferability is demonstrated by experiments on a real robot, for which we synthesize tracking controllers.
\end{abstract}


%
\IEEEpeerreviewmaketitle

\input{sections/Introduction.tex}
\input{sections/Preliminaries.tex}
\input{sections/Method_A.tex}
\input{sections/Method_B.tex}

\input{sections/Method_C.tex}

\input{sections/Evaluation.tex}

\input{sections/Discussion.tex}

\input{sections/Conclusion.tex}
\input{sections/Appendix.tex}

\bibliographystyle{IEEEtran}
\bibliography{references,references_bastian}

%

\begin{IEEEbiography}[{\includegraphics[width=1in,height=1.25in,clip,keepaspectratio]{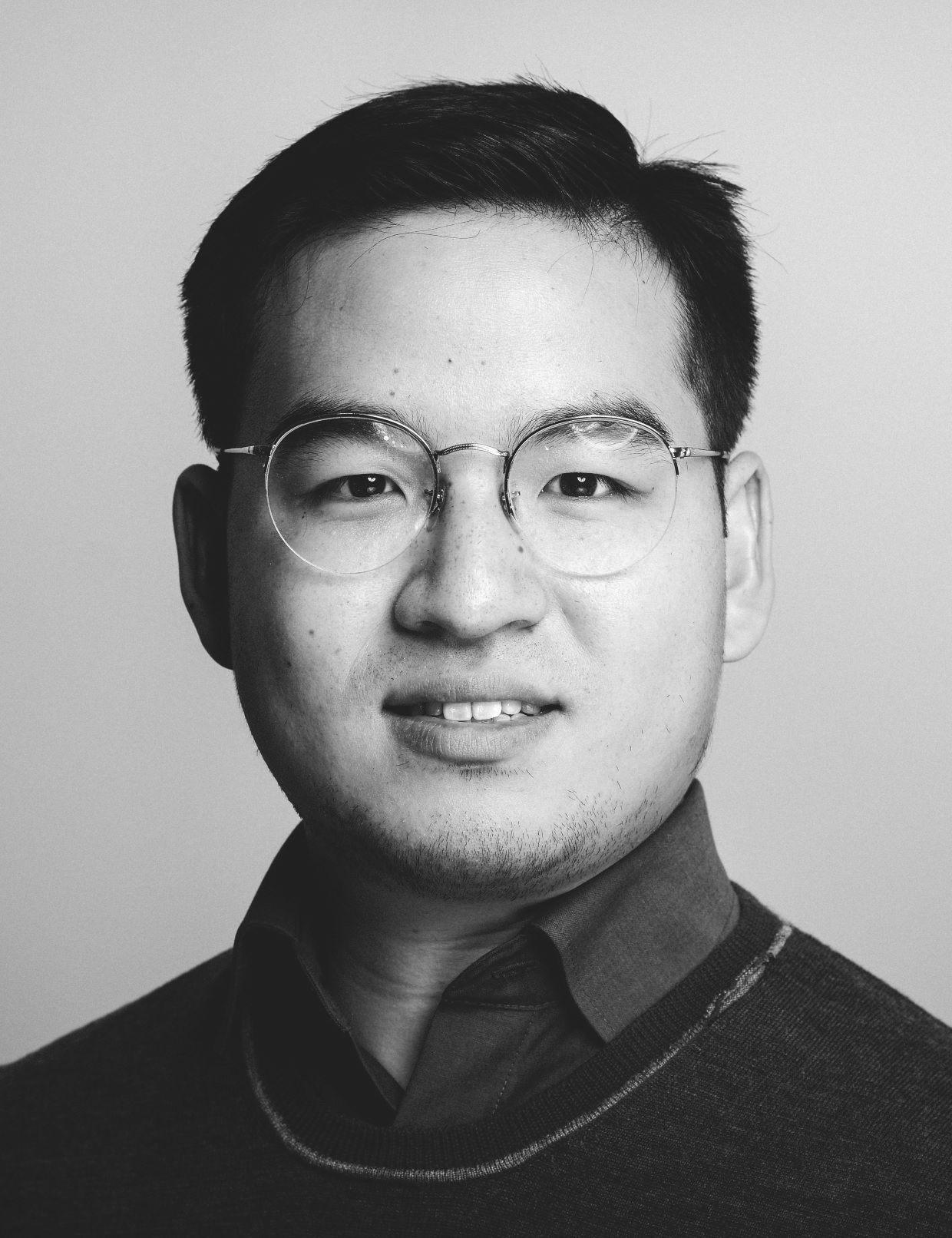}}]{Stefan B. Liu}
received a B.S. degree in mechatronics,
and an M.S. degree in robotics from the
Technical University of Munich (TUM), Germany,
in 2015 and 2017, respectively. He is currently
pursuing a Ph.D. degree at the Cyber-Physical Systems
Group of the TUM Department of Informatics.
His research interest includes formal methods in
robotics, physical human-robot interaction, modeling
and identification, and modular robots.
\end{IEEEbiography}

\begin{IEEEbiography}[{\includegraphics[width=1in,height=1.25in,clip,keepaspectratio]{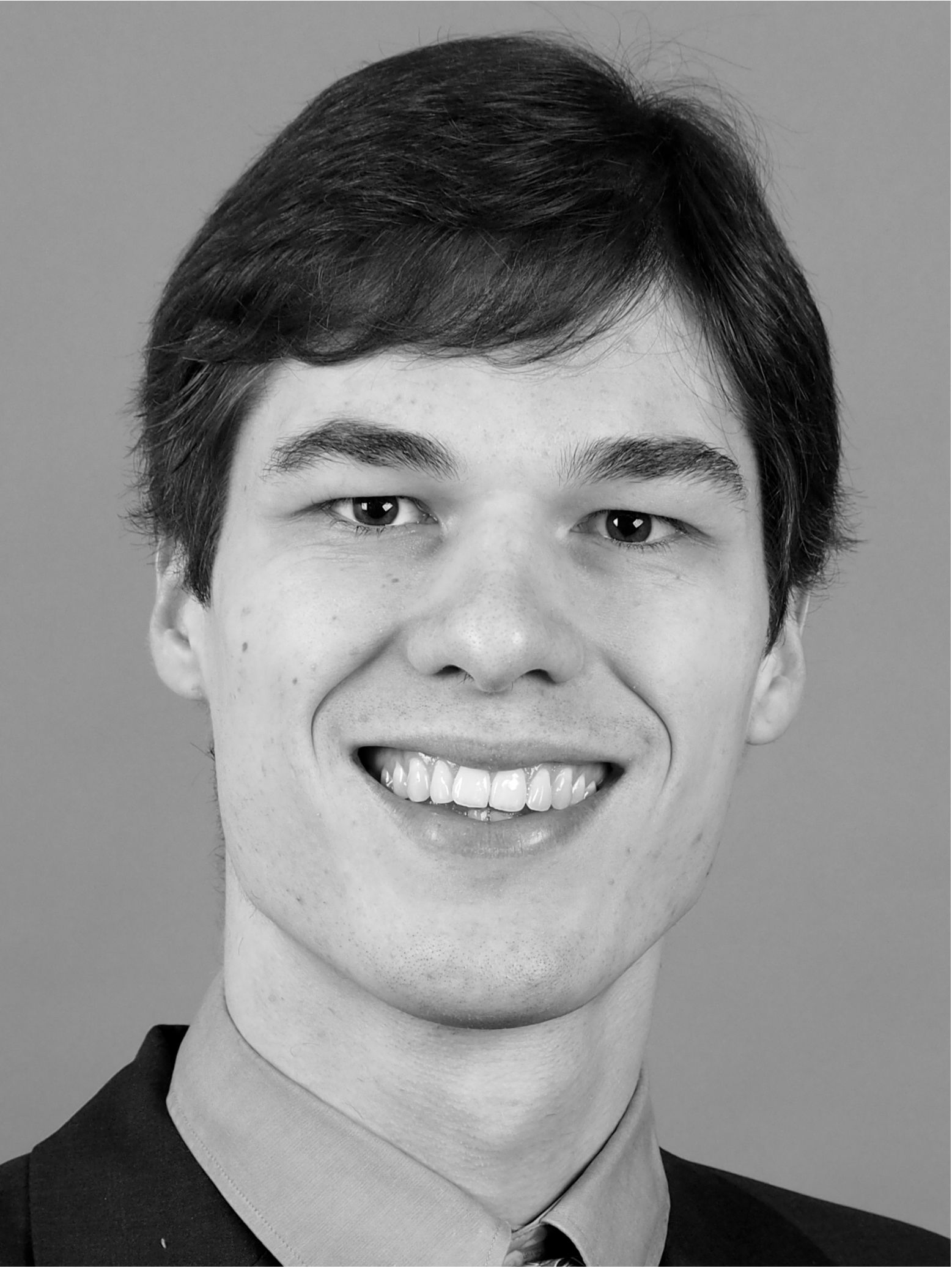}}]{Bastian Sch\"urmann}
 received a Bachelor of Science in Electrical and Computer Engineering from  Techni\-sche Universit\"at Kaiserslautern, Germany, in 2012; a Master of Science in Electrical Engineering from the University of California, Los Angeles, USA, in 2014; a Master of Science in Engineering Cybernetics from Universit\"at Stuttgart, Germany, in 2015; and a Ph.D. in Informatics from Technische Universit\"at M\"unchen in 2022. In 2018, he was a visiting student researcher at the California Institute of Technology. His research focuses on combining control theory, reachability analysis, and optimization.
\end{IEEEbiography}

\begin{IEEEbiography}[{\includegraphics[width=1in,height=1.25in,clip,keepaspectratio]{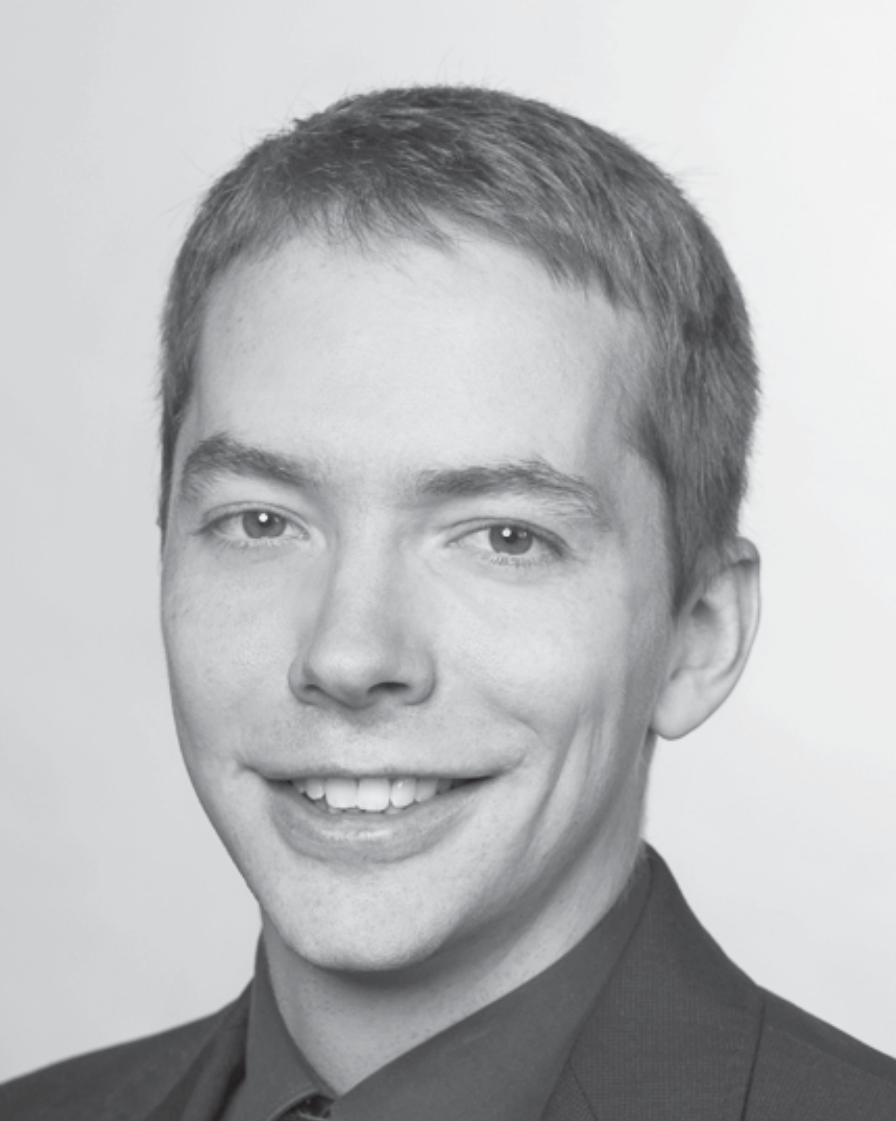}}]{Matthias Althoff} 
	is an Associate Professor in computer science at the Technical University of Munich, Germany. He received his Diploma Engineering Degree in Mechanical Engineering in 2005 and his Ph.D. in Electrical Engineering in 2010, both from the Technical University of Munich, Germany. From 2010 to 2012, he was a postdoctoral researcher at Carnegie Mellon University, Pittsburgh, USA, and from 2012 to 2013, he was an assistant professor at the Ilmenau University of Technology, Germany. His research interests include formal verification of continuous and hybrid systems, reachability analysis, planning algorithms, nonlinear control, robotics, automated vehicles, and power systems.
\end{IEEEbiography}


\end{document}

%% file: sections/Introduction.tex

\section{Introduction}\label{sec_introduction}

Guaranteeing and optimizing control performance has been a challenge for the robust control of robots for a long time (e.g., see the surveys in  \cite{Abdallah1991,Sage1999}). One of the reasons is that models of robots and their controllers do not consider certain effects: 1) rigid-body models of robots do not consider flexible joints and links; 2) some model parameters are falsely assumed to be constant, e.g., some friction parameters in robots depend on  load and temperature, which are not accounted for in standard models; and 3) control limitations, such as finite motor capabilities, finite sampling
time, measurement errors, delays, noise within circuit boards, etc., are typically not modeled. Due to these and other reasons, an identified model can never exhibit exactly the same behavior as the real system.

We propose a novel formal synthesis framework that uses \textit{reachability analysis} \cite{Althoff2020} to optimize the controller and provide formal guarantees for robotic systems. Reachability analysis allows us to formally bound all possible behaviors, making it possible to decide whether a given specification is always met.

Our main challenge is how to correctly identify models such that the guarantees obtained for these transfer to the corresponding real robot. We will make use of the \textit{reachset conformance} relation \cite{Roehm2019}, which means that the reachable sets of the model must contain all possible behaviors of the real robot. Broadly speaking: if a property can be guaranteed for a conservative model, then we can guarantee the same property for the real system (a formal explanation will be provided in Sec. \ref{sec_preliminaries}). In this paper, we combine reachset-conformant identification with controller synthesis in a single optimization problem that simultaneously finds the optimal model and controller. 
Obviously, if one is interested in only identifying a reachset-conformant model or only finding a controller for a given model of a robot, our approach is also applicable. 

This paper focuses on the synthesis of tracking controllers for feedback-linearized robots, but is applicable to all linear systems. The software, as well as the scripts to replicate our experimental results, can be obtained from Code Ocean\footnote{\url{https://doi.org/10.24433/CO.1635335.v1}}.

\subsection{Literature overview}\label{sec_survey}
We divide our review of relevant works into three parts: robust control, formal synthesis, and model identification.
\subsubsection{Robust control}

Previous robustness analyses of feedback-linearizing robot controllers, many of which are surveyed in \cite{Abdallah1991} and \cite{Sage1999}, assume that system uncertainties originate from model errors, which can be considered additive nonlinear disturbances in the feedback-linearized model. For instance, the nonlinear disturbance representation helps to prove general uniform ultimate boundedness (UUB) for a computed torque controller in \cite{Qu1991}. In \cite[Section 8.5.3]{Siciliano2009a}, a robust controller is proposed, where UUB is shown by bounding the mass matrix and other nonlinear terms of the robot dynamics. The approach in \cite{Zenieh1997} presents a control scheme for robots that achieves a desired tracking error with a pre-specified convergence rate. Generally, in previous works, UUB is only shown through Lyapunov's theorem, which can be very tedious. In contrast, we quantitatively model the additive disturbances as an uncertain set and show UUB directly by computing the reachable tracking error of a robot using standard algorithms for reachability analysis~\cite{Althoff2020}. These algorithms also make it possible to incorporate sampling times, measurement errors, and delays---all of which influence the final tracking error.

$\mathcal{H}_\infty$-synthesis (e.g., in \cite{Kim2015,Makarov2016}) is a method that optimally designs robot controllers that minimize an $\mathcal{H}_\infty$-norm, which captures disturbance effects expressed in the frequency domain. However, $H_\infty$-synthesis does not provide any guarantees with respect to input constraints. Similarly, the linear quadratic regulator (LQR) is an optimization-based approach, which has robustness properties  \cite{Lin1998} but fails to consider constraints (more details in Sec. \ref{sec_evaluation}).

A well-known type of controller ensuring the satisfaction of state and input constraints despite the presence of disturbances is tube-based model predictive control (MPC). There, an optimization algorithm iteratively optimizes a reference trajectory over a moving horizon while a feedback controller keeps the system in a tube around the reference trajectory. For linear systems, the computation of the reference trajectory and the control invariant set of the tube can be decoupled due to the superposition principle \cite{Mayne2005,LANGSON2004125,rakovic2012parameterized,rakovic2012homothetic}, while for nonlinear systems, this becomes more complex. Still, a number of approaches exist for nonlinear systems, e.g., \cite{rubagotti2011robust,Magni2003Robust,Mayne2011Tube,singh2017robust}.
Other ways to ensure the satisfaction of constraints are to embed an invariance controller \cite{WOLFF200537,kimmel2014invariance} or use control barrier functions \cite{ames2016control,wieland2007constructive}.
In contrast to tube-based MPC, our approach meets the specification for the real robot and not just its model. In addition, our approach does not require finding a Lyapunov function, as required for most tube-based MPC approaches.

\subsubsection{Formal synthesis}
Formal controller synthesis is a research area with many recent results in robotics; we refer to \cite{Kress-Gazit2018} for an overview. The idea is to compute a controller which formally guarantees the satisfaction of complex specifications. Many of the control approaches mentioned in \cite{Kress-Gazit2018} focus on high-level planning with little focus on uncertainty, while our method formally synthesizes low-level controllers, where uncertainty plays a larger role.

Many formally correct controllers are realized as abstraction-based controllers \cite{kloetzer2008fully,zamani2012symbolic,decastro2015synthesis,fainekos2009temporal,girard2012controller,kress2009temporal,liu2013synthesis,Liu2016Finite,pola2007symbolic,raman2015reactive,rungger2013specification}, which satisfy rich specifications such as temporal logic expressions. By discretizing the state and input space, they obtain a finite state abstraction of the system so that they can use techniques from automata theory to synthesize controllers. The necessity to discretize the state space leads to an exponential computational complexity with respect to the number of continuous state variables, which restricts the application to lower-dimensional systems. Some works try to avoid this problem by not abstracting the whole state space, e.g., see \cite{zamani2015symbolic,wolff2016optimal,DeCastro2016Nonlinear}. In contrast to these papers, we avoid discretizing the state space and directly compute the reachable set of the dynamic system, which scales polynomially with the number of state variables \cite{Althoff2020}.

Instead of abstracting the whole state space, other approaches compute safe motion primitives for mobile robots, i.e., short trajectory pieces with a corresponding controller that keeps the system in predefined sets. By computing many motion primitives and storing them in a maneuver automaton, they can be used with a discrete online planner, which only needs to find a suitable concatenation of motion primitives \cite{saha2014automated,sanfelice2008hybrid}. There are different methods to compute these motion primitives, e.g., using LQR trees \cite{tedrake2010lqr,majumdar2016funnel}, or by combining optimization with reachability analysis \cite{Schuermann2017a,Schuermann2017c,Schuermann2017b}. For robotic systems, such as manipulators, precomputing motion primitives would be infeasible since the number of required motion primitives scales exponentially with the number of states and inputs. Instead, our goal is to provide guarantees for the tracking error independently from the desired motion. 

Other techniques, such as interval arithmetics \cite{Calzolari2020a} or Hamilton-Jacobi reachability \cite{Chen2021}, can also be used to compute and ensure the tracking error bounds of dynamical systems given known disturbances. In the next few paragraphs, we will review techniques that help us if disturbances are unknown.

\subsubsection{Identification of model uncertainties}

Uncertainties can be generally categorized as stochastic and set-based uncertainties formulated in the frequency or time domain \cite{VanDenHof1995}.  A discussion of  uncertainties in the frequency domain for robust control can be found in \cite{Douma2005}. Stochastic aspects of model uncertainty are treated in large detail in \cite{Ljung1999}. For instance, in \cite{Santolaria2013}, the stochastic uncertainty of robot kinematics is identified through Monte Carlo sampling. Since we focus on providing guarantees, we will discuss set-based uncertainties in the time domain.

Formal synthesis requires models that enclose the behavior of real systems. This is also called the \textit{model conformance} relation and has been treated in-depth in \cite{Roehm2019}.
Most literature on set-based identification is based on finding a \textit{simulation relation} since it allows a transfer of, e.g., temporal logic properties for the entire state space. The model can be a coarse-grained abstraction of the state space into a discrete automaton (e.g., for the navigation of mobile robots \cite{Kress-Gazit2018}) or differential inclusions \cite{Chen2019a,Sadraddini2018}. The paper in \cite{Chen2019a} identifies a linear system with non-determinism such that all state measurements are within a polytopic reachable set. The paper in \cite{Sadraddini2018} identifies piece-wise affine models using mixed-integer linear programming, also establishing a simulation relation between measured states with hyperrectangular reachable sets. In contrast to these works, we use zonotopes, which have a special structure that allows us to reduce the identification to a linear problem.

However, if a system is high-dimensional, but only a few outputs are relevant for synthesis, then the simulation relation can be too restrictive and conservative. Thus, \textit{trace} and \textit{reachset conformance} have been proposed to relax the formal relation only to the output of a system \cite{Roehm2019}. In \cite{Schurmann2018}, the authors apply trace conformance by reconstructing disturbance traces for a real autonomous vehicle. The set of non-deterministic disturbances is then taken as the outer bounds of all disturbance traces. \textit{Reachset conformance}, on the other hand, is a further relaxation that only requires that the output traces of a system must be within the reachable set of the model. The main advantage is that we can handle sensor noise and arbitrary disturbances, which is not possible for trace conformance since this would create infinitely many possible behaviors, resulting in a more flexible model-order reduction \cite{Althoff2012a} or even applying black-box identification methods \cite{Wang2021}. 
For transferring safety properties, reachset conformance is sufficient \cite{Roehm2019}.

Our previous work on the reachset conformance of robot manipulators, on which this paper is based, can be found in \cite{Liu2018,Giusti2021}. Our work in \cite{Liu2018} aims to identify the uncertain sets of a forward dynamical model, while here, we identify a feedback-linearized robot model. In \cite{Giusti2021}, a reachset-conformant inverse dynamical robot model is identified. In these works, we have not combined reachset-conformant identification with controller synthesis.

The identification of conformant parameter sets can also be viewed as a synthesis problem. The authors in \cite{Dang2019,Batt2007} are able to incorporate additional model knowledge as temporal logic constraints to improve identification results.

The main criterion for the identification of parameter sets is usually the size of their range. However, small uncertainties do not necessarily lead to good robust control, and large model errors do not necessarily lead to bad control performance, as \cite{Skelton1989} has pointed out. Therein lies the motivation for \textit{identification for control}, in which the model uncertainties are determined in a way that is optimal for the control goal \cite{VanDenHof1995}. Our framework builds upon these ideas to formulate controller synthesis and model identification as a unified optimization problem, where they share a common cost function.

Notably, set-membership identification \cite{Vicino1996,Milanese2004,Kieffer2006,Bravo2006,Ramdani2005} has certain similarities to our approach because it is also a set-based method. There, the goal is to identify the true parameter of a system by reducing the feasible solution set as much as possible. This is different from reachset-conformant identification, where the goal is to model the parameter set large enough to ensure reachset conformance. Parameters obtained from set-membership identification are generally not reachset conformant and cannot be used for our robust control framework.

\subsection{Structure of this paper}
This paper is structured as follows: in Sec. \ref{sec_preliminaries}, we provide preliminaries on zonotopes and on the reachability analysis of uncertain linear systems. Our combined controller synthesis and reachset-conformant identification framework is presented in Sec. \ref{sec:methods}. We address the application of these methods to the tracking control problem of robots in Sec. \ref{sec_evaluation} and conclude this paper in Sec.~\ref{sec:conclusion}.

%% file: sections/Preliminaries.tex
\section{Preliminaries and Problem Statement}\label{sec_preliminaries}
We first introduce preliminaries on set operations and subsequently describe the control problem.

\subsection{Preliminaries on set operations}
We denote sets using calligraphic letters (e.g., $\mathcal{A}$), matrices using upper case letters (e.g., $A$), vectors using $\vec{\cdot}$, and scalar values using lower case letters (e.g., $a$). 
To represent sets, we mainly use zonotopes.
\begin{definition}[Zonotope]\label{def:zonotopeG}
	A zonotope $\mathcal{Z}$ is defined by a center $\vec{c}$ and a generator matrix $G$ of proper dimension, where $\vec{g}^{(h)}$ is its $h$-th column:
	\begin{align*}
		\mathcal{Z} &= (\vec{c},G) :=\left\lbrace \vec{x}=\vec{c} + \sum_{h=1}^{s}\beta_h \vec{g}^{(h)} \Bigg\vert \beta_h \in [-1,1] \right\rbrace.
	\end{align*}
\end{definition}
\noindent A $\theta$-dimensional zonotope $\mathcal{Z}$ with $s$ generators can also be described by an intersection of $2{s \choose \theta-1}$ half-spaces. 
\begin{proposition}[H-representation of a zonotope \cite{Althoff2010a}]\label{def:zonotopeH}
	The half-space representation of a zonotope is $\{ \vec{y}\, | N \vec{y} \leq \vec{d} \}$,
	\begin{equation*}
		N = \begin{bmatrix}
			N^+ \\ -N^+
		\end{bmatrix},\quad
		\vec{d} = \begin{bmatrix}
			\vec{d}^+\\ \vec{d}^-
		\end{bmatrix},
	\end{equation*}
	where each row of $N$ and $\vec{d}$ contains the normal vectors and distances of a half-space, respectively. The direction of each normal vector is computed from a reduced generator matrix $G^{\langle\gamma,\dots,\eta\rangle}$, where $\gamma,\dots,\eta$ are the $s-\theta+1$ indices of the generators that have been removed from $G$. The $j$-th row of $N^+$, where $j \in 1..{s \choose \theta-1}$, is 
	\begin{gather}
		\vec{n}_j^+ = \nX (G^{\langle\gamma,\dots,\eta\rangle})/ \|\nX (G^{\langle\gamma,\dots,\eta\rangle})\|_2 \label{eq:problem:normalVector},\\
		\nX(H) := [\dots,  (-1)^{i+1}\det(H^{[i]}), \dots]^T,
	\end{gather}
	where $H^{[i]}$ means that the $i$-th row of $H$ is removed, and the $j$-th row of $\vec{d}^+$ and $\vec{d}^-$ are
	\begin{gather}
		d_j^+ = \vec{n}_j^{+T} \, \vec{c} + \Delta d_j, \qquad
		d_j^- = -\vec{n}_j^{+T} \, \vec{c} + \Delta d_j \label{eq:problem:dj},\\
		\Delta d_j = \sum_{h=1}^{s}|\vec{n}_j^{+T} \, g^{(h)}|.\label{eq:problem:ddj}
	\end{gather}
\end{proposition}

Many operations on zonotopes can be exactly and efficiently computed \cite{Althoff2020}. Let us define the Minkowski sum of sets  as $\mathcal{A} \oplus \mathcal{B} = \lbrace \vec{a} + \vec{b} \mid \vec{a} \in \mathcal{A}, \vec{b} \in \mathcal{B}\rbrace$. For zonotopes, the following propositions hold:
\begin{proposition}[Minkowski sum of zonotopes \cite{Girard2005}]\label{prop:minkSum}
	Zonotopes are closed under Minkowski sum:
	\begin{align*}
		\mathcal{Z}_1 \oplus \mathcal{Z}_2 =  (\vec{c}_1,G_1) \oplus  (\vec{c}_2,G_2) =  (\vec{c}_1+\vec{c}_2,[G_1,G_2]).
	\end{align*}
\end{proposition}

\begin{proposition}[Linear transformation of zonotopes \cite{Girard2005}]\label{prop:linTrans}
	Zonotopes are closed under linear transformation: 
	\begin{equation*}
		A \mathcal{Z} =  (A\vec{c},AG).
	\end{equation*}
\end{proposition}

To reason about the size of a zonotope, we introduce a norm that is defined based on the edge lengths of its interval hull. Alternative norms can be found in \cite{Gassmann2020}.

\begin{proposition}[Interval hull of zonotopes \cite{Girard2005}]\label{def:intervalHull}
	The interval hull $\mathcal{I}(\mathcal{Z}) = [\vec{i}^-,\vec{i}^+]$, where $\vec{i}^-$ is the left bound and $\vec{i}^+$ is the right bound, is the smallest interval enclosing a set $\mathcal{Z} = (\vec{c},[\dots,\vec{g}^{(h)},\dots])$, where
	\begin{gather*}
		\vec{i}^- = \vec{c} - \vec{\delta}, \qquad \vec{i}^+ = \vec{c} + \vec{\delta}, \qquad \vec{\delta} = \sum_{h=1}^{p} |\vec{g}^{(h)}|.
	\end{gather*}
\end{proposition}

\begin{definition}[Norm of zonotopes]\label{def:zonotopeNorm}
	We define the norm of a zonotope as the sum of each element of $\vec{\delta}$, which represents the size of the interval hull: 
	\begin{equation*}
		\|\mathcal{Z}\| := \sum_{i=1}^\theta |\delta_i|.
	\end{equation*}
\end{definition}

Next, we introduce reachable sets for linear systems. Since robots are commonly measured and controlled by computers, we restrict our discussion to discrete time. We use the notation $a[k]$ to express the value of $a$ at time $k\Delta t$, where $k \in \lbrace 0,1,\dots\rbrace$ and $\Delta t$ is the sampling time. Discrete-time linear systems are defined by the following difference and output equations:
\begin{align} \label{eq:problem:discreteSystem}
\begin{split}
\vec{x}[k+1] &= A \vec{x}[k] + B \vec{u}[k] + \vec{w}[k],\\ 
\vec{y}[k] &= C \vec{x}[k] + D \vec{u}[k] + \vec{v}[k],
\end{split}
\end{align}
where $A,B,C,D$ are matrices of proper dimension, $\vec{x}[k]$ is the state, $\vec{y}[k]$ is the output, $\vec{u}[k] \in \mathcal{U}$ is the control input constrained by $\mathcal{U}$, and $\vec{w}[k] \in \mathcal{W} = (\vec{c}_W,G'_W\diag(\vec{\alpha}_W))$ and $\vec{v}[k] \in \mathcal{V} = (\vec{c}_V,G'_V\diag{(\vec{\alpha}_V}))$ are the disturbances sensor noise, respectively, bounded by appropriate zonotopes to capture the errors of the nominal model. The operator $\diag(.)$ returns a matrix where the elements of the input vector are on the diagonal. Subsequently, vectors $\vec{\alpha}_W$ and $\vec{\alpha}_V$ are variables that scale the length of each generator of $\mathcal{W}$ and $\mathcal{V}$, respectively.  

Reachable sets are defined as the set of all possible outputs of a system, given a set of initial states and the set of all possible inputs. The reachable set of \eqref{eq:problem:discreteSystem} after one time step is computed through a set-based evaluation of the difference and output equations in \eqref{eq:problem:discreteSystem}:
\begin{multline}
\mathcal{R}[k+1] = C \left(A \mathcal{X}[k] \oplus B \vec{u}[k] \oplus \mathcal{W}\right) 
\oplus D \vec{u}[k+1] \oplus \mathcal{V}, \label{eq:problem:reachableSetStep}
\end{multline}
where $\mathcal{X}[k]$ is the current set of states. Given an initial set $\mathcal{X}[0]$, the reachable set after $k$ time steps can be computed by recursively applying \eqref{eq:problem:reachableSetStep}:
\begin{multline}
\mathcal{R}[k+1] = C \left(  A^{k+1} \mathcal{X}[0] \oplus \sum_{i=0}^{k} A^i B \vec{u}[i]
\oplus \bigoplus_{i=0}^{k}A^{i}\mathcal{W} \right) \\ 
 \oplus D \vec{u}[k+1] \oplus \mathcal{V}. \label{eq:problem:reachableSet}
\end{multline}
When using zonotopes, the above computation is exact since \eqref{eq:problem:reachableSet} only involves Minkowski sums and linear transformations.

\subsection{Plant model and reachset conformance}\label{sec:plantModel}

In this subsection, we discuss the model of our use case. Because our method applies to linear systems and the robot dynamics are nonlinear in general, we implement an internal \textit{feedback linearization} in the robot.
Let us derive the plant model by regarding the following rigid-body dynamics of a robot \cite[Sec. 2.2]{Sage1999}:

\begin{equation}\label{eq:problem:robotDynamics}
	M(\vec{q})\ddot{\vec{q}} + \vec{\psi}(\vec{q},\dot{\vec{q}})= \vec{\tau},
\end{equation}
where $\vec{q}$ is the vector of joint positions, $\vec{\tau}$ is the vector of joint torques, $M$ is the mass matrix, and $\vec{\psi}$ contains the Coriolis, centripetal, gravity, and friction forces.
The feedback linearization technique \cite{Siciliano2009a} applies a control torque
\begin{equation}\label{eq:problem:feedbackLinearizationTorque}
	{\vec{\tau} = M(\vec{q})\vec{u}_{r} + \vec{\psi}(\vec{q},\dot{\vec{q}})}
\end{equation}
to \eqref{eq:problem:robotDynamics}; for the rigid-body dynamics, this results in linear dynamical systems that are decoupled for each joint $i$:
\begin{equation}\label{eq:problem:feedbackLinearization}
	\ddot{q}_i = u_{r,i},
\end{equation}
where $u_{r,i}$ is the plant input for the feedback-linearized robot with rigid-body dynamics.
In the discretized state-space model for one robot joint, we additionally consider that both the input and the output are delayed by one sampling instant.
Let us denote the linear dynamics by the subscript $r$ (for \emph{robot}):
\begin{align}
\begin{split}
	\vec{x}_r[k+1] &= \begin{bmatrix}	0 & 1 & 0 & 0 \\ 0 & 1 & \Delta t &  \frac{\Delta t^2}{2} \\ 0 & 0 & 1 & \Delta t \\ 0 & 0 & 0 & 0	\end{bmatrix} \vec{x}_r[k] + \begin{bmatrix} 0 \\ 0 \\ 0 \\ 1 \end{bmatrix} u_r[k], \\
	y_r(t) &= \begin{bmatrix}	1 & 0 & 0 & 0\end{bmatrix} \vec{x}_r[k], 
	\end{split}\label{eq:problem:robotLinearDynamics}
\end{align}
where $\vec{x}_r[k] = [q[k-1],q[k],\dot{q}[k],u_r[k-1]]^T$ is the state, and $y_r$ is the measured joint position.

The dynamics of a real robot, however, will never be exactly as in \eqref{eq:problem:robotLinearDynamics} because 1) the rigid-body assumption has limitations, e.g., there are flexible parts in the system, 2) the inertial parameters used in the feedback linearization in \eqref{eq:problem:feedbackLinearizationTorque} are usually not accurate, and 3) measurement errors affect the feedback linearization. To capture the uncertainties of the robot, we add the following uncertainties: a time-varying additive disturbance $\vec{w}_{{p},i}[k] \in \mathcal{W}_{{p},i} \subseteq \mathbb{R}^3$, and a measurement error $v_{{p},i}[k] \in\mathcal{V}_{{p},i} \subseteq \mathbb{R}$, and an additional constant disturbance state $d$, where $\dot{d} = 0$. The full model is denoted the subscript $p$ (for \emph{plant}) and is fully described in Appendix \ref{sec:appendix1}.

The next definitions specify the data we require to test for reachset conformance.

\begin{definition}[Test case]\label{def:testCase}
	Let $k^*\in \mathbb{N}_0$. A test case is a tuple $(y_p[0],...,y_p[k^*], u_p[0],...,u_p[k^*], \vec{x}_p[0])$ of output measurements $y_p[k]$, control inputs $u_p[k]$, and the initial state $\vec{x}_p[0]$.
\end{definition}
\noindent To account for disturbances at any point in time, we should generate \textit{sequential} test cases (defined subsequently) to have as many initial states as possible and to maximize the number of test cases from one recording.
\begin{definition}[Sequential test cases]\label{def:seqtestCase}
	From one recording, we generate multiple test cases, where the state of each time step can be the start of a new test case. Sequential test cases are denoted by a superscripted index. The following relation holds for sequential test cases: 
	\begin{gather*}
		\vec{y}_p^{(m+1)}[k] = \vec{y}_p^{(m)}[k+1], \\
		u_p^{(m+1)}[k] = u_p^{(m)}[k+1].
	\end{gather*}
\end{definition}

Finally, we establish reachset conformance \cite[Sec. 3.5]{Roehm2019} by testing the real system.

\begin{definition}[Reachset conformance testing]\label{def:reachsetConformance}
	Given are a plant model and $M$ test cases of a real system. The model is reachset conformant for the sampling instants $k \in \lbrace 0,...,k^*\rbrace$ if, for each test case, the measurement of the real system is enclosed in the corresponding reachable set of the model:
	\begin{equation*}
	\forall m  \, \forall k: \vec{y}_p^{(m)}[k] \in \mathcal{R}_p^{(m)}[k],
	\end{equation*}
	where $\vec{y}_p^{(m)}[k]$ is the measured output and $\mathcal{R}_p^{(m)}[k]$ is computed using \eqref{eq:problem:reachableSet} considering $\vec{x}^{(m)}[0]$ and $u^{(m)}[k]$.
\end{definition}
\noindent We call finding of unknown parameters of the plant model, such that Def. \ref{def:reachsetConformance} is fulfilled, \textit{reachset-conformant identification}.

\subsection{Problem statement}\label{sec:synthesisProblem}

Now, let us discuss the problem at hand. Our goal is to synthesize an optimal closed-loop system given a linear plant model, while the disturbance sets have unknown parameters to be identified. The control goal is for the output of the closed-loop system $\vec{y}_\mathrm{cl}:= [\hat{q}, \dot{\hat{q}}]^T$ to track a reference output $\vec{y}_\mathrm{ref} := [q_d, \dot{q}_d]$ containing the desired position and velocity. The observed variables $\hat{q}, \dot{\hat{q}}$ have been chosen for $\vec{y}_\mathrm{cl}$ since the robot velocity is usually not measurable, so an observer \cite{Liu2022} is recommended.

For the closed-loop system, we select a parameterizable linear feedback controller and a parameterizable linear observer such that the closed-loop system is also linear, and its reachable set can be computed using \eqref{eq:problem:reachableSet}. Furthermore, we can include input feedforward signals $u_\mathrm{ff}$ that are added to the plant input, e.g., a desired acceleration $u_\mathrm{ff} := \ddot{q}_d$. In Sec.~\ref{sec_evaluation}, we demonstrate two different closed-loop systems with unknown parameters.

Next, we specify the optimization problem for the combined controller synthesis and reachset-conformant identification. Subsequently, we define the two main reachable sets considered in our controller synthesis:

\begin{definition}[Reachable tracking error]\label{def:Re}
$\mathcal{R}_e$ is a reachable set that encloses all tracking errors of the closed-loop system, such that
\begin{equation*}
	\vec{y}_\mathrm{cl} \in \vec{y}_\mathrm{ref} \oplus \mathcal{R}_e.
\end{equation*}
\end{definition}
\begin{definition}[Reachable input]
$\mathcal{R}_u$ is the reachable set of all plant inputs $u_p$ in the closed-loop dynamics. A controller is considered safe if the reachable input is within the allowed set $\mathcal{U}_p$, such that $\mathcal{R}_u \subseteq \mathcal{U}_p$.
\end{definition}
\noindent The computation of $\mathcal{R}_e$ and $\mathcal{R}_u$ are explained in Sec. \ref{sec:method:cost}.
As a cost function, we choose the norm of the reachable tracking error $\mathcal{R}_e$. The variables are the unknown controller and observer parameters, as well as $\vec{c}_{W_p},\vec{c}_{V_p},\vec{\alpha}_{W_p},\vec{\alpha}_{V_p}$ from the zonotopic disturbances of the plant model. These variables are aggregated into a parameter vector $\vec{p} \in \mathcal{P}$, where $\mathcal{P}$ is a user-defined search space. 
The optimization problem has two constraints:
\begin{itemize}
	\item the plant model shall be reachset conformant (Def. \ref{def:reachsetConformance}),
	\item the plant input is constrained so that we never exceed the allowed motor torques of the robot,
\end{itemize}
and the optimization problem is formulated as:
\begin{subequations}\label{eq:problem:optimProblem}
	\begin{alignat}{2}
	&\!\min_{\vec{p} \in \mathcal{P}}        &\qquad& \|\mathcal{R}_{e}(\vec{p})\|, \label{eq:problem:cost}\\
	&\text{subject to} &      &  \forall m \, \forall k \in [0,k^*]: \, {\vec{y}}_p^{(m)}[k] \in \mathcal{R}^{(m)}_p(\vec{p})[k],  \label{eq:problem:constraintConformance}\\
	&				     &      &  \mathcal{R}_{u}(\vec{p}) \subseteq \mathcal{U}_p, \label{eq:problem:constraintInput}
	\end{alignat}
\end{subequations}
where all computed reachable sets depend on $\vec{p}$. The optimization problem is defined for each robot axis $i \in \{1..n\}$, but the set of allowed inputs $\mathcal{U}_{p,i}$ for each axis are derived from the allowed joint torque and depend on the axis configuration. Given the feedback linearization in \eqref{eq:problem:feedbackLinearizationTorque}, the allowed set of inputs $\mathcal{U}_p = \mathcal{U}_{p,1} \times ... \times \mathcal{U}_{p,n}$ must satisfy the torque limits:
\begin{gather}\label{eq:problem:deriveInputConstraint}
	\mathcal{T} \supseteq M(\mathcal{Q})\mathcal{U}_p \oplus \vec{\psi}(\mathcal{Q},d\mathcal{Q}),
\end{gather}
where $\mathcal{T}$ is the set of allowed torques, and $\mathcal{Q},d\mathcal{Q}$ are the sets of allowed positions and velocities of the robot. Since \eqref{eq:problem:deriveInputConstraint} is nonlinear, we recommend Taylor models \cite{Makino2003,Althoff2018} as a set representation to prove the above statement because the precision of Taylor models in approximating nonlinear functions can be set arbitrarily high.

The main advantage of this combined approach is that all parameters are synthesized for the same goal, while an approach with separate goals would lead to sub-optimal models.
Notice, however, that a standalone reachset-conformant identification can be derived from the above problem by leaving out \eqref{eq:problem:constraintInput} and switching to any other cost function, e.g., a prediction error as demonstrated in \cite{Liu2018,Giusti2021,Liu2021}. Also, by removing \eqref{eq:problem:constraintConformance}, we arrive at the standalone controller synthesis problem proposed in \cite{Schuermann2017b}. 

%% file: sections/Method_A.tex
\section{Combined controller synthesis and reachset-conformant identification}\label{sec:methods}
This section describes how to solve \eqref{eq:problem:optimProblem}. In Sec. \ref{sec:method:cost}, we first explain the computation of the reachable tracking error $\mathcal{R}_e$ and the reachable input $\mathcal{R}_u$. In Sec. \ref{sec:method:conformance}, we derive a linear formulation of reachset conformance \eqref{eq:problem:constraintConformance}, which reduces the complexity of the constraint evaluation to a linear inequality check. In Sec. \ref{sec:method:implementation}, we discuss the need to solve \eqref{eq:problem:optimProblem} iteratively and cover the computational aspects in Sec.~\ref{sec:method:computationalAspects}.

\subsection{Computing the reachable tracking error and input}\label{sec:method:cost}

Often in robotics, the desired position and velocity may not be known in advance, e.g., when using online trajectory generation. Therefore, our aim is to solve \eqref{eq:problem:optimProblem} independently from the reference. Nevertheless, we shall restrict the desired acceleration by a set $u_\mathrm{ff} \in \mathcal{U}_\mathrm{ff}$ to disallow unbounded feedforward inputs. 
To later extract both $\mathcal{R}_{e}$ and $\mathcal{R}_{u}$ as a projection \cite[Sec.~2.1]{Althoff2015} of the reachable set of the closed-loop system, we augment its output by the plant input $u_p$; the new output is denoted by a tilde: $\tilde{\vec{y}}_\mathrm{cl} := [\vec{y}_\mathrm{cl},u_p]^T$.

Similar to \cite{Schuermann2017b}, we use the superposition principle for linear systems to divide the reachable set of the closed-loop system into two parts: a set $\tilde{\mathcal{R}}_\mathrm{cl,e}$  that is only dependent on the disturbances $\mathcal{W}_p$ and $\mathcal{V}_p$, and a vector $\tilde{\vec{y}}_\mathrm{cl,ref}$ that is only dependent on the reference $\vec{y}_\mathrm{ref}$ and the feedforward $u_\mathrm{ff}$, such that the final reachable set is $\tilde{\vec{y}}_\mathrm{cl,ref}[k] \oplus \tilde{\mathcal{R}}_\mathrm{cl,e}[k]$. 

\begin{figure}
	\includegraphics[width=0.9\columnwidth]{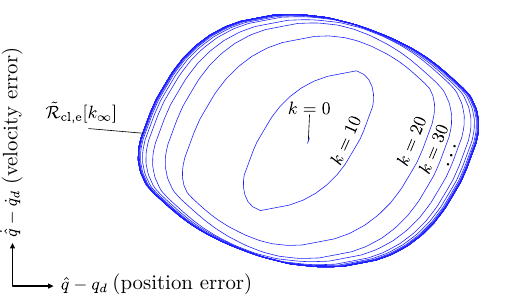}
	\caption{Computation of the reachable tracking error, which is a projection of the converged set $\tilde{\mathcal{R}}_\mathrm{cl,e}[k_\infty]$.}
	\label{fig:method:reachSetError}
\end{figure}

The set $\tilde{\mathcal{R}}_\mathrm{cl,e}$ is computed using \eqref{eq:problem:reachableSet} by setting $\vec{y}_\mathrm{ref} = 0$ and $u_\mathrm{ff}=0$. If the system is stable, then $\tilde{\mathcal{R}}_\mathrm{cl,e}[k]$ will converge to an invariant set $\tilde{\mathcal{R}}_\mathrm{cl,e}[k_\infty]$ \cite{Gruber2020}, i.e., $\tilde{\mathcal{R}}_\mathrm{cl,e}[k_\infty+1] \subseteq \tilde{\mathcal{R}}_\mathrm{cl,e}[k_\infty]$.
In practice, this convergence might not happen due to numerical issues; therefore, we implement \cite[Alg.~2]{Gruber2020}, which computes $\tilde{\mathcal{R}}_\mathrm{cl,e}[k_\infty]$ from an arbitrarily small and an arbitrarily large $\mathcal{X}(0)$ until they converge to a final set with a tolerance criterion that is chosen to be arbitrarily small.
Thus, the computed $\tilde{\mathcal{R}}_\mathrm{cl,e}[k_\infty]$ is a positive invariant set \cite{Gruber2020} of both the tracking error and the plant input. 
An example of $\tilde{\mathcal{R}}_\mathrm{cl,e}[k]$ converging to $\tilde{\mathcal{R}}_\mathrm{cl,e}[k_\infty]$ is shown in Fig.  \ref{fig:method:reachSetError}.

If $\tilde{\vec{y}}_\mathrm{cl,ref}[k] = [\vec{y}_\mathrm{ref}[k],u_\mathrm{ff}]^T$, then the sets for the reachable tracking error and the reachable input are given by the following projections; since the reachable input $\mathcal{R}_u$ should also contain $u_\mathrm{ff}$, we add the bounded set $\mathcal{U}_\mathrm{ff}$:
\begin{align*}
	\mathcal{R}_e &:= \begin{bmatrix} I_{i} & 0_{i\times j} \end{bmatrix} \tilde{\mathcal{R}}_\mathrm{cl,e}[k_\infty],\\
	\mathcal{R}_u &:= \begin{bmatrix} 0_{j\times i}  & I_{j}  \end{bmatrix} \tilde{\mathcal{R}}_\mathrm{cl,e}[k_\infty] \oplus \mathcal{U}_\mathrm{ff},
\end{align*}
where $I_{i\times i}$ is an identity matrix with dimension $i$, $0_{i\times j}$ is a matrix of zeros with $i$ rows and $j$ columns, $i$ is the dimension of $\vec{y}_\mathrm{cl}$, and $j$ is the dimension of $u_p$. 
 However, we note that $\vec{y}_\mathrm{ref}[k]$ is not always equal to $\vec{y}_\mathrm{cl,ref}[k]$; extensions considering the remaining error can be found in Appendix \ref{sec:appendix2}.

%% file: sections/Method_B.tex

\subsection{Reachset conformance as a set of linear inequalities}\label{sec:method:conformance}

If constraint \eqref{eq:problem:constraintConformance} was naively implemented, the reachable set $\mathcal{R}^{(m)}_p$ would have to be computed for each test case. However, we will show  that \eqref{eq:problem:constraintConformance} can be reduced to a set of linear inequalities depending on $\vec{\xi} = [\vec{c}_{W_p},\vec{c}_{V_p},\vec{\alpha}_{W_p},\vec{\alpha}_{V_p}]^T$. For the remainder of this subsection, all variables refer to the plant model, and the subscript $p$ is omitted for ease of notation.

The first simplification is to combine all test cases as described subsequently to check reachset conformance by a single reachability analysis. 
Let us define $\vec{y}^{(m)}_{*}[k]$ as the nominal solution of \eqref{eq:problem:reachableSet} for the plant without the disturbance sets $\mathcal{W}$ and $\mathcal{V}$, and consider $\vec{x}^{(m)}[0]$ as the initial state.
To make \eqref{eq:problem:constraintConformance} independent of each test case, 
we subtract the nominal solution from both $\vec{y}^{(m)}[k]$ and the reachable set $\mathcal{R}^{(m)}[k]$:
\begin{align}
	\vec{y}_{a}^{(m)}[k] &:= \vec{y}^{(m)}[k] - \vec{y}^{(m)}_{*}[k],\label{eq:method:ya}\\
	\mathcal{R}_a[k] &:= \mathcal{R}^{(m)}[k] - \vec{y}^{(m)}_{*}[k] \overset{\eqref{eq:problem:reachableSet}}{=} \bigoplus_{i=0}^{k-1} C A^{i}\mathcal{W}
	\oplus \mathcal{V}, \label{eq:method:Ra}
\end{align}
where $\vec{y}_{a}^{(m)}[k]$ is the deviation of the real behavior from the nominal one, and $\mathcal{R}_a[k]$ is now independent of the input and the initial state. Therefore, for linear systems, the following statement is equal to \eqref{eq:problem:constraintConformance}:
\begin{equation}\label{eq:methods:distConstr}
	 \forall k \in \{0..k^*\}: \, \bigcup_m \{\vec{y}_{a}^{(m)}[k]\} \subseteq \mathcal{R}_{a}[k],
\end{equation}
where the left side is the union of all trajectories deviating from the nominal behavior. Since $\mathcal{W}$ and $\mathcal{V}$ are zonotopes, we can apply propositions \ref{prop:minkSum} and \ref{prop:linTrans} to derive that the center and generator matrix of $\mathcal{R}_a[k] = (\vec{c}_k,G_k)$ are
\begin{gather}
	\vec{c}_k := \begin{bmatrix} \sum_{i=0}^{k-1}E_i & 1 \end{bmatrix} \begin{bmatrix} \vec{c}_W \\ \vec{c}_V \end{bmatrix}, \quad E_i = C A^{i},\label{eq:methods:reachA}\\
	G_k := \begin{bmatrix} E_0 G_W & \dots & E_{k-1} G_W & G_V \end{bmatrix}. \label{eq:method:Gk}
\end{gather}

The second simplification is to formulate \eqref{eq:methods:distConstr} as a set of linear inequalities by using the H-representation of $\mathcal{R}_a$ (see Proposition \ref{def:zonotopeH}): if all $\vec{y}_a^{(m)}[k]$ satisfy all the half-space inequalities of $\mathcal{R}_a[k]$ for all $k$, then \eqref{eq:methods:distConstr} follows, and the model is reachset conformant. As we will show in the following theorem, the half-space inequalities for $\mathcal{R}_a$ are not only linear in $\vec{y}_a^{(m)}$, but they are also linear in $\vec{\xi}$. The directions of the half-space normal vectors do not depend on $\vec{\xi}$, but our optimization in \eqref{eq:problem:optimProblem} is rather varying the distance of each half-space from the measured outputs. The number of test cases can be arbitrarily large since we will use the measurement with the largest deviation from the nominal behavior.
\begin{theorem}\label{theorem:linearConstraint}
	The constraint \eqref{eq:problem:constraintConformance} for the reachset conformance of linear systems is linear in  $\vec{\xi} = [\vec{c}_W,\vec{c}_V,\vec{\alpha}_W,\vec{\alpha}_V]^T$:
	\begin{equation}\label{eq:methods:linearConstraint}
		\forall k \in \left[0,k^*\right]:\, \max_m(N_k y_a^{(m)}[k]) \leq D_k \vec{\xi},
	\end{equation}
	where $N_k = [N_k^+,-N_k^+]^T$, and $D_k = [D_k^+, D_k^-]$. The $j$-th row of $N_k^+$, where $j \in 1..{s \choose \theta-1}$, is a normal vector of the H-representation of $\mathcal{R}_a[k]$ and independent from $\xi$:
	\begin{align*}
		\vec{n}_{j,k}^+ &= \nX ({G'_k}^{\langle\gamma,\dots,\eta\rangle})^T/ \|\nX ({G'_k}^{\langle\gamma,\dots,\eta\rangle})\|_2,
	\end{align*}
	and the $j$-th row of $D_k^+$ and $D_k^-$ are defined as
	\begin{align*}
		\begin{split}
			\vec{d}_{j,k}^+ &= 
			\Big[\begin{matrix} \sum_{i=0}^{k-1} \vec{n}_{j,k}^+ E_i, & \vec{n}_{j,k}^+, \end{matrix} \\
			&\qquad\qquad \begin{matrix} \sum_{i=0}^{k-1} |\vec{n}_{j,k}^+ E_i  G'_W|,& |\vec{n}_{j,k}^+  G'_V| \end{matrix}\Big],
		\end{split} \\
		\begin{split}
		\vec{d}_{j,k}^- &= 
		\Big[\begin{matrix} -\sum_{i=0}^{k-1} \vec{n}_{j,k}^+ E_i, & -\vec{n}_{j,k}^+,  \end{matrix} \\
		&\qquad\qquad \begin{matrix} \sum_{i=0}^{k-1} |\vec{n}_{j,k}^+ E_i  G'_W|,& |\vec{n}_{j,k}^+ G'_V| \end{matrix}\Big].
		\end{split}
	\end{align*}	
\end{theorem}
\begin{proof}
	We demonstrate that the normal vectors of the H-representation of any zonotope~$(\vec{c}, G' \diag(\vec{\alpha}))$ are independent from $\vec{\alpha}$. The numerator of $\vec{n}_{j}^+$ is (see \eqref{eq:problem:normalVector}):
	\begin{align*}
		\nX &(G\diag(\vec{\alpha})) =\\
		&=  [\dots,  (-1)^{i+1}\det(G^{[i]}\diag(\vec{\alpha})), \dots]^T\\
		&=  \det(\diag(\vec{\alpha}))[\dots,  (-1)^{i+1}\det(G^{[i]}), \dots]^T\\
		&= \det(\diag(\vec{\alpha})) \cdot \nX (G) = \rho \nX (G),
	\end{align*}
	and since all elements of $\vec{\alpha}$ are positive, we infer $\rho > 0$, and the denominator of $\vec{n}_{j}^+$ is
	\begin{align*}
		\big\| \rho \nX (G)\big\|_2 = \rho \|  \nX (G)\|_2,
	\end{align*}
	such that $\vec{\alpha}$ cancels out from the definition of $\vec{n}_{j}^+$  in \eqref{eq:problem:normalVector}. 
	Next, we show that $\vec{d}_{j,k}^+$ and $\vec{d}_{j,k}^-$ can be derived from applying the definition of $G_k$ in \eqref{eq:method:Gk} to \eqref{eq:problem:ddj} in Proposition \ref{def:zonotopeH}, considering $\vec{1}$ as a vector of ones:
	\begin{align*}
		&\Delta d_{j,k} = \begin{bmatrix} | \vec{n}_{j,k}^+ E_0 G_W | & \dots & | \vec{n}_{j,k}^+ E_{k-1} G_W |& | \vec{n}_{j,k}^+G_V |\end{bmatrix} \, \vec{1} \\
		&= \begin{bmatrix} |\vec{n}_{j,k}^+E_0 G'_W| & \dots & |\vec{n}_{j,k}^+E_{k-1} G'_W  |& |\vec{n}_{j,k}^+G'_V| \end{bmatrix}\begin{bmatrix} \alpha_W \\ \vdots \\ \alpha_W \\ \alpha_V \end{bmatrix} \\
		&= \begin{bmatrix} \sum_{i=0}^{k-1} |\vec{n}_{j,k}^+E_i G'_W| & |\vec{n}_{j,k}^+G'_V| \end{bmatrix}\begin{bmatrix} \alpha_W \\ \alpha_V \end{bmatrix}.
	\end{align*}
	The first two elements of $\vec{d}_{j,k}^+$ and $\vec{d}_{j,k}^-$ directly follow from \eqref{eq:problem:dj}, which are linear in the zonotope center $\vec{c}$, so that $\vec{d}_{j}^+$ and $\vec{d}_{j}^-$ in Proposition \ref{def:zonotopeH} are linear in $\vec{\xi}$ when $G_k$ is applied.
\end{proof}

One problem which we could encounter is that the number of constraints is $2{p \choose n-1}$ and exponentially increases with $k$ since $p$ (the number of generators of $G_k$ in \eqref{eq:method:Gk}) grows for each time step. The following corollary can be used for a conservative approximation of the linear inequality, which reduces the number of constraints, yet guarantees reachset conformance for $k^* \rightarrow \infty$. This is achieved by making the estimated states of the plant $\vec{x}^{(m)}$ conformant to the reachable set of the plant states. As the following proof will show, this reduces \eqref{eq:problem:constraintConformance} to a simple inclusion check for $\mathcal{W}$ and $\mathcal{V}$.

\begin{corollary}\label{lemma:simpleLinear}
	Let us consider sequential test cases. Then, a linear system is reachset conformant for $k^* \rightarrow \infty$, if
	\begin{gather} 
		   \bigcup_{m \in \{1..M\}} \{\vec{x}_{a}^{(m)}[1]\} \subseteq \mathcal{W}, \label{eq:method:globalReachConfW} \\
		 \bigcup_{m \in \{1..M\}} \{\vec{y}^{(m)}[0] - C\vec{x}^{(m)}[0] - D\vec{u}^{(m)}[0]\}\subseteq \mathcal{V}, \label{eq:method:globalReachConfV}
	\end{gather}
	where 
	\begin{equation}\label{eq:def:xa}
		\vec{x}_{a}^{(m)}[k] := \vec{x}^{(m)}[k] - \vec{x}^{(m)}_*[k]
	\end{equation}
	is the deviation from the nominal state: $\vec{x}^{(m)}_*[0] = \vec{x}^{(m)}[0]$ and $\vec{x}^{(m)}_*[1] = A\vec{x}^{(m)}[0] + Bu^{(m)}[0]$.
\end{corollary}
\begin{proof}
	We first rewrite the original problem before we perform the actual proof. Let us define a set $\mathcal{R}_{x,a}$ as the reachable set of states of $\vec{x}_{a}$, such that
	\begin{gather} 
		\mathcal{R}_{x,a}[k+1] = A \mathcal{R}_{x,a}[k]\oplus \mathcal{W}, \quad \mathcal{R}_{x,a}[0] = \vec{x}_a[0] = \vec{0}. \label{eq:method:Xa}
	\end{gather}
	The reachable output in \eqref{eq:method:Ra} can thus be rewritten as
	\begin{equation*}
		\mathcal{R}_a[k]  = C \mathcal{R}_{x,a}[k]\oplus \mathcal{V},
	\end{equation*}
	and the definition of reachset conformance in \eqref{eq:methods:distConstr} can be rewritten as
	\begin{gather}
		\forall k \in \mathbb{N}_0: \,\bigcup_m \{C\vec{x}_{a}^{(m)}[k] + v[k]\} \subseteq C\mathcal{R}_{x,a}[k] \oplus\mathcal{V}.\label{eq:method:CXpV}
	\end{gather} 
	We can derive reachset conformance by proving that the summands of \eqref{eq:method:CXpV} are conformant. Since \eqref{eq:method:globalReachConfV} is given,
	\begin{gather}
		\bigcup_m v^{(m)}[k] \subseteq \mathcal{V}
	\end{gather} 
	for any $k$. Next, we show $\forall k: C\vec{x}_a^{(m)}[k] \in C\mathcal{R}_{x,a}[k]$. Here, we prove 
	\begin{equation}
		\vec{x}_a^{(m)}[k] \in\mathcal{R}_{x,a}[k], \quad k\rightarrow\infty
	\end{equation}
by induction, when \eqref{eq:method:globalReachConfW} is given. 
Using $u^{(m+k)}[0] = u^{(m)}[k]$ from Def. \ref{def:seqtestCase}, we derive
	\begin{align}\label{eq:method:zwischen}
	\begin{split}
		&\vec{x}^{(m+k)}_*[1] - \vec{x}_*^{(m)}[k+1] \\
		 &= A\vec{x}^{(m+k)}[0] + Bu^{(m+k)}[0] - A\vec{x}_*^{(m)}[k] - Bu^{(m)}[k] \\
		 &= A(\vec{x}^{(m)}[k] - \vec{x}_*^{(m)}[k]) \\
		 &= A\vec{x}_a^{(m)}[k].
	\end{split}
	\end{align}
	Since \eqref{eq:method:globalReachConfW} is given, the base case for $k=1$ holds:
	\begin{equation*}
	\forall m: \, \vec{x}_a^{(m)}[1] \in A\mathcal{R}_{x,a}[0] \oplus \mathcal{W} = \mathcal{X}_a[1] \label{eq:method:conf1},
	\end{equation*}
	because $\mathcal{X}_a[0] = \vec{0}$. 
	Now we apply the induction step $k+1$:
	\begin{gather*}
	\vec{x}_a^{(m)}[k+1] \overset{\eqref{eq:def:xa}}{=}  \vec{x}^{(m+k)}[1] - \vec{x}_*^{(m)}[k+1]\\
	=  \vec{x}^{(m+k)}_a[1] + \vec{x}^{(m+k)}_*[1] - \vec{x}_*^{(m)}[k+1]\\
	\overset{\eqref{eq:method:zwischen}}{=} A\vec{x}_a^{(m)}[k] +  \vec{x}^{(m+k)}_a[1] 
	\overset{\eqref{eq:method:globalReachConfW}}{\in}  A\vec{x}_a^{(m)}[k]  \oplus \mathcal{W} \\
	\overset{\text{induction hypothesis}}{\subseteq} A\mathcal{R}_{x,a}[k] \oplus \mathcal{W} \overset{\eqref{eq:method:Xa}}{=}  \mathcal{R}_{x,a}[k+1]. \label{eq:method:conf2}
	\end{gather*}
\end{proof}
\begin{remark}
The H-representation of $\mathcal{W}$ and $\mathcal{V}$ can also be used to formulate the constraints in the corollary as linear inequalities. The proof is similar to Theorem \ref{theorem:linearConstraint}.
\end{remark}
\begin{remark}
	Using Corollary 1 to identify the disturbances is generally more conservative than using Theorem 1: if the column rank of $C$ is not full, then checking $\vec{x}_a \in \mathcal{R}_{x,a}$ is more strict than checking $\vec{y}_a \in \mathcal{R}_a = C\mathcal{R}_{x,a} \oplus \mathcal{V}$ because $C\mathcal{R}_{x,a}$ is a projected set. Another explanation is that the conformance of states constitutes a simulation relation \cite[Sec. 3.3]{Roehm2019}, which entails the conformance of outputs. 
\end{remark}
\begin{remark}
	In practice, a threshold exists where any larger $k^*$ does not affect the results of Theorem 1 anymore. This threshold can be found by testing the synthesis with increasing $k^*$. For Corollary 1, this step is not required.
\end{remark}

%% file: sections/Method_C.tex

\subsection{Iterative synthesis} \label{sec:method:implementation}

\begin{figure}
	\includegraphics[width=0.9\columnwidth]{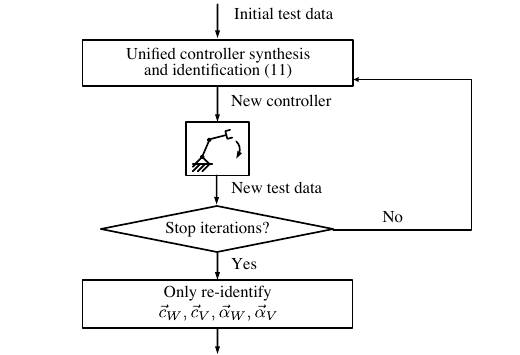}
	\caption{Iterative procedure for simultaneous reachability-based identification and control synthesis.}
	\label{fig:methods:iterative}
\end{figure}

As has been demonstrated in \cite{Skelton1989}, the error of the nominal plant model can change depending on the chosen controller parameters, e.g., our nominal model does not consider flexible elements, which could lead to vibrations when controller parameters are ill-chosen. Since we use $\mathcal{W}_p$ and $\mathcal{V}_p$ to enclose the model errors, these sets, therefore, could also change depending on the controller parameters.

When we solve \eqref{eq:problem:optimProblem}, a new set of controller parameters are proposed. We, therefore, need an iterative approach (see Fig. \ref{fig:methods:iterative}) that adjusts the sets $\mathcal{W}_p$ and $\mathcal{V}_p$ based on re-testing the real robot, which in turn influences the controller synthesis again. Similar to previous concepts in identification for control \cite{VanDenHof1995}, we propose the following iterations:
\begin{enumerate}
	\item Given an initial set of test data obtained using an initial controller, we compute the linear constraint using Theorem 1 and solve \eqref{eq:problem:cost}--\eqref{eq:problem:constraintInput} to synthesize an optimal controller.
	\item Using the new controller, we repeat the tests on the real robot and obtain a new set of test data.
	\item Repeat step 1 with the new data to synthesize a new controller. If no further iteration is desired, then we perform a re-identification of the disturbances only, i.e., solving \eqref{eq:problem:cost}--\eqref{eq:problem:constraintInput} without changing the controller, to obtain the final result.
\end{enumerate}

\subsection{Computational aspects}\label{sec:method:computationalAspects}
The optimization problem in \eqref{eq:problem:optimProblem} is posed as a nonlinear program with a nonlinear cost function \eqref{eq:problem:cost}. If a solution exists, we are able to check reachset conformance \eqref{eq:problem:constraintConformance} and satisfy input constraints \eqref{eq:problem:constraintInput}. However, we cannot guarantee convergence to a globally optimal solution; we can only expect to obtain a local optimum. Nevertheless, practical tuning rules can be helpful in improving convergence, e.g., consider a static feedback controller \cite[eq.~8.58]{Siciliano2009a} that we will consider in Sec. \ref{sec_evaluation}: $u_p = \ddot{q}_d + k_p (q_d - q) + k_d(\dot{q}_d - \dot{q})$. By replacing the  parameters $k_p=\omega^2$ and $k_d=2\zeta\omega$ with the natural frequency $\omega$ and damping ratio $\zeta$, the convergence improved. Such tuning rules were initially developed for manual tuning to converge faster to an optimal solution and can obviously also serve as hints to improve convergence for our automatic approach.

We cannot provide concrete complexity bounds for nonlinear programming since no bounds exist for them. Nevertheless, let us give an idea of the complexity of the different evaluations. The cost \eqref{eq:problem:cost} and the constraint function \eqref{eq:problem:constraintInput} mainly involve computing reachable sets and some algebraic operations on the resulting zonotopes, which together have a complexity of $\mathcal{O}(n^3)$ \cite{Schuermann2021a}, where $n$ is the number of states. The conformance constraints in \eqref{eq:problem:constraintConformance} can be efficiently evaluated since they are linear inequalities. Checking the constraint $\mathcal{R}_u \subseteq \mathcal{U}_p$ in \eqref{eq:problem:constraintInput} requires only checking if a zonotope is inside a polytope, which can also be efficiently computed \cite[Lemma 2]{Schuermann2021a}.

%% file: sections/Evaluation.tex
\section{Experiments on a 6-axis robot manipulator}\label{sec_evaluation}

\begin{figure}
	\includegraphics[width=\columnwidth]{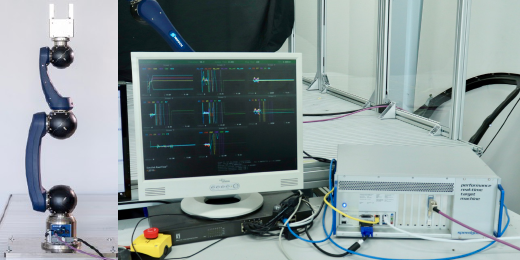}
	\caption{The testbed consists of a Schunk LWA-4P 6-DOF robot manipulator and a controller running on Simulink Real-Time.}
	\label{fig:experiment:setup}
\end{figure}

In this section, we show the results of applying our combined controller synthesis and reachset-conformant identification to a real 6-axis robot manipulator (see Fig. \ref{fig:experiment:setup}). In the first experiment in Sec. \ref{sec:eval:separate}, we work out the benefits of using the combined approach in comparison to separate identification and synthesis. In the second experiment in Sec. \ref{sec:eval:LQR}, we compare our method against the linear-quadratic-Gaussian control (LQG). In the third experiment in Sec. \ref{sec:eval:ESOvsHG}, we demonstrate how our method makes it possible to compare  the guarantees of different controllers.

The data for testing reachset conformance were obtained from the real robot running closed-loop trapezoidal and polynomial trajectories\footnote{A video showing the initial tests, and the code for reproducing all experiments are provided within the supplementary materials.} with random target positions, velocities, and accelerations up to $\ddot{q}_d \in \mathcal{U}_\mathrm{ref} = [-2,2]$ rad/$s^2$. The total duration of the dataset is 33 minutes and 20 seconds. Each sampling instant is considered a starting point of a new test case, resulting in 497,880 test cases for each robot joint. Other test selection methods (e.g., \cite{Deshmukh2017,Roehm2019}) can be used to find test cases that explore edge scenarios more effectively; however, a basic approach---such as random testing---may already be sufficient. An inherent problem with testing will always be that there are cases that are not covered by the tested trajectories. In addition, changes to the robot dynamics can happen that are also not covered by the test cases. We propose to implement \eqref{eq:problem:constraintConformance} as an online conformance monitor that detects non-conformant measurements, transitions the system to a safe stop, and repeats identification for this new test case. If the resulting new disturbance violates the input constraint in \eqref{eq:problem:constraintInput}, the controller synthesis needs to be repeated. 

The time horizon for reachset conformance has been selected to be $k^* = 125$. At a sampling time $\Delta t = 0.004$ s, this amounts to $0.5$ seconds. Because $y_p$ is one-dimensional, this amounts to 252 conformance constraints (two half-spaces per time step, including $k=0$). 
To check whether a selected $\mathcal{U}_p$ satisfies the allowed set of joint torques $\mathcal{T} := [-\vec{\tau}_\mathrm{max},\vec{\tau}_\mathrm{max}]$, we set $\mathcal{Q} := [-\vec{q}_\mathrm{max},\vec{q}_\mathrm{max}]$ and $d\mathcal{Q} := [-\dot{\vec{q}}_\mathrm{max},\dot{\vec{q}}_\mathrm{max}]$ and evaluate \eqref{eq:problem:deriveInputConstraint} using tenth-order Taylor models \cite{Althoff2018}. The values can be seen in Table \ref{tab:eval:inputConstraint}. To avoid the wrapping effect, which accumulates approximation errors, we split $\mathcal{Q}$ into four intervals and evaluate \eqref{eq:problem:deriveInputConstraint} for each interval combination.

\begin{table}
	\caption{Deriving $\mathcal{U}_p$ from specified robot limits using \eqref{eq:problem:deriveInputConstraint}}
	\label{tab:eval:inputConstraint}
	\setlength\tabcolsep{6pt}
	\renewcommand{\arraystretch}{1.25}
	\begin{center}
		\begin{tabular}{ r r r r  r}
			\hline
			Axis&$\vec{\tau}_\mathrm{max}$ & $\vec{q}_\mathrm{max}$ & $\dot{\vec{q}}_\mathrm{max}$ & $\mathcal{U}_p$ satisfying \eqref{eq:problem:deriveInputConstraint}\\ \hline
			$1$ & $160$ Nm & $140^\circ$ & $0.7$ rad/s & $[-26,26]$ rad/$s^2$\\
        	$2$ & $160$ Nm & $ 45^\circ$ & $0.7$ rad/s & $[-26,26]$ rad/$s^2$\\ 
        	$3$ & $160$ Nm & $100^\circ$ & $0.7$ rad/s & $[-26,26]$ rad/$s^2$\\  
        	$4$ & $160$ Nm & $140^\circ$ & $0.7$ rad/s & $[-26,26]$ rad/$s^2$\\ 
       		$5$ &  $40$ Nm & $ 80^\circ$ & $0.7$ rad/s & $[-100,100]$ rad/$s^2$\\  
        	$6$ &  $40$ Nm & $140^\circ$ & $0.7$ rad/s & $[-100,100]$ rad/$s^2$\\ 
			\hline
		\end{tabular}
	\end{center}
\end{table}

\subsection{Combined vs. separate identification and synthesis} \label{sec:eval:separate}

\begin{figure}
	\includegraphics[width=\columnwidth]{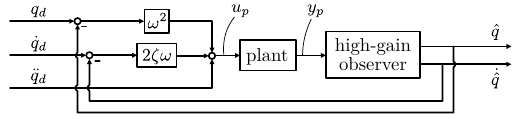}
	\caption{The control loop considered in Sec. \ref{sec:eval:separate} tracks the reference $\vec{y}_\mathrm{ref}= [q_d,\dot{q}_d]$ and has an input feedforward $u_\mathrm{ff} = \ddot{q}_d$. A high-gain observer \cite{Nicosia1993a} is used to observe the position $q$ and velocity $\dot{q}$, which are used in the feedback control.}
	\label{fig:eval:controllerSynthesis1}
\end{figure}

\begin{figure}
	\includegraphics[width=\columnwidth]{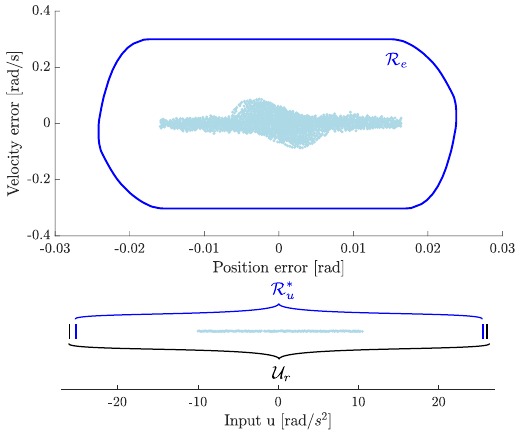}
	\caption{After solving \eqref{eq:problem:optimProblem}, both the computed reachable tracking error $\mathcal{R}_e$ and the computed reachable input $\mathcal{R}_u$ enclose their measured counterparts, while $\mathcal{R}_u$ satisfies the input constraint.}
	\label{fig:eval:SF}
\end{figure}

In the first experiment, we compare our combined approach against a separate approach, where a reachset-conformant model is identified before the controller synthesis. The controller-observer structure for this experiment is depicted in Fig. \ref{fig:eval:controllerSynthesis1} and is chosen as follows: a high-gain observer \cite{Nicosia1993a} uses the plant output $y_p$ to estimate $\hat{q}$ and $\dot{\hat{q}}$:
\begin{gather}
\label{eq:eval:robotObserver}
	\begin{bmatrix} \dot{\hat{q}} \\ \ddot{\hat{q}} \end{bmatrix} = \begin{bmatrix}	-h_1/\epsilon & 1 \\ -h_2/\epsilon^2 & 0	\end{bmatrix} \begin{bmatrix} \hat{q} \\ \dot{\hat{q}} \end{bmatrix} + \begin{bmatrix} h_1/\epsilon \\ h_2/\epsilon^2 \end{bmatrix} y_p,
\end{gather}
where $h_1 = 15,h_2 = 30$, and $\epsilon:=0.01$ are the gains. To discretize the observer, we use the bilinear transformation discussed in \cite{Busawon2017b}. As the controller, we consider a static feedback one \cite[eq.~8.58]{Siciliano2009a}:
\begin{equation}\label{eq:eval:staticfeedback}
	u_p = \ddot{q}_d + \omega^2 (q_d - \hat{q}) + 2\zeta\omega(\dot{q}_d - \dot{\hat{q}}),
\end{equation}
where $\omega$ and $\zeta$ are the parameters to be optimized. For the combined approach, we set $\vec{p} = [\omega,\zeta, \alpha_{W_p,1},\alpha_{W_p,2},\alpha_{V_p}]^T$ and solve \eqref{eq:problem:optimProblem} for two iterations. The final result can be seen in Tab. \ref{tab:eval:SF}, and we plot $\mathcal{R}_e$ and $\mathcal{R}_u$ for the first robot axis in Fig. \ref{fig:eval:SF}. Our combined approach returned feasible solutions for all six axes. The uncertainties for axes 5 and 6 are larger than others, mainly due to the inaccuracy of the feedback linearization for these axes. As Fig. \ref{fig:eval:SF} shows, our reachable sets correctly predict the real tracking errors and the real inputs.

\begin{table}
	\caption{Optimally synthesized Static Feedback ($\omega,\zeta$) and identified~Model uncertainties ($\alpha_{W,1},\alpha_{W,2},\alpha_{V,1}$)}
	\label{tab:eval:SF}
	\setlength\tabcolsep{6pt}
	\renewcommand{\arraystretch}{1.25}
	\begin{center}
		\begin{tabular}{ r r r r r r r }
			\hline
			Axis&$\|\mathcal{R}_e\|$ & $\omega$ & $\zeta$ & $\alpha_{W_p,1}$ & $\alpha_{W_p,2}$ & $\alpha_{V_p,1}$ \\ \hline
			$1$ & $0.62$ & $24.40$ & $0.96$ & $0.0348$ & $3.59$ & $3.49\cdot 10^{-5}$\\
			$2$ & $0.70$ & $24.39$ & $0.96$ & $0.0442$ & $3.82$ & $3.49\cdot 10^{-5}$\\ 
			$3$ & $0.58$ & $23.06$ & $0.91$ & $0.0381$ & $2.93$ & $3.49\cdot 10^{-5}$\\ 
			$4$ & $0.58$ & $23.81$ & $0.96$ & $0.0363$ & $3.00$ & $3.49\cdot 10^{-5}$\\
			$5$ & $0.91$ & $24.71$ & $1.00$ & $0.0142$ & $7.72$ & $3.49\cdot 10^{-5}$\\
			$6$ & $2.39$ & $25.23$ & $1.00$ & $0.0000$ & $23.28$ & $2.80\cdot 10^{-5}$\\ 
			\hline
		\end{tabular}
	\end{center}
\end{table}

\begin{figure*}
	\includegraphics[width=180mm]{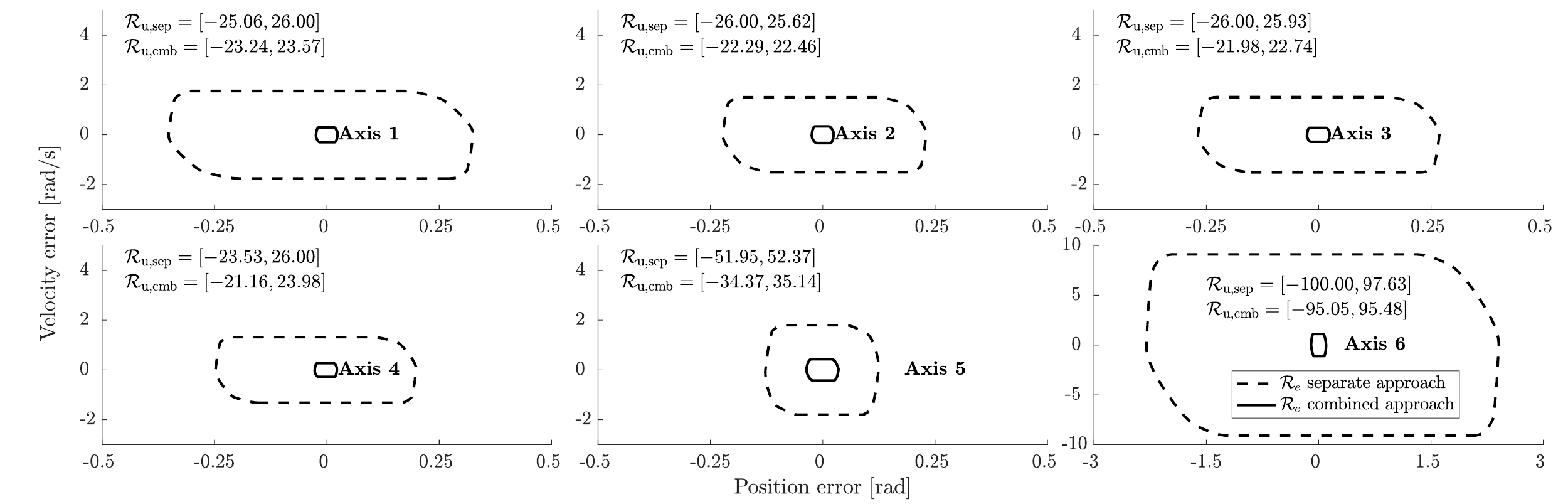}
	\caption{Comparing $\mathcal{R}_e$ and $\mathcal{R}_u$ for the separate approach against the combined approach for identification and synthesis. A separate identification may lead to suboptimal $\mathcal{W}_p$ and $\mathcal{V}_p$, such that the closed-loop reachable sets become unnecessarily large. In the separate approach, the controller synthesis converged to smaller gains, e.g., $\omega = 5.9$ for axis 1 to satisfy the input constraint; compared to the combined approach, where for axis 1, $\omega = 24.4$.}
	\label{fig:eval:separateVScombined}
\end{figure*}

For the separate approach, we first identify a reachset-conformant model by solving an optimization problem, where  $\|\mathcal{W}_p\| + \|\mathcal{V}_p\|$ is set as the cost function and \eqref{eq:problem:constraintConformance} is set as the constraint function, and $\alpha_{W_p,1},\alpha_{W_p,2},\alpha_{V_p}$ are the parameters. For the subsequent controller synthesis, we set \eqref{eq:problem:cost} as the cost, \eqref{eq:problem:constraintInput} as the constraint, and $\omega,\zeta$ as the parameters. The plots in Fig. \ref{fig:eval:separateVScombined} show that the separate approach leads to a significantly larger reachable set $\mathcal{R}_e$, although the identified values $\alpha_{W_p,1} = 0.75$ and $\alpha_{W_p,2} = 0$ for axis 1 lead to a smaller value of $\|\mathcal{W}_p\|$ than the values identified in the combined approach $\alpha_{W_p,1} = 0.0348$ and $\alpha_{W_p,2}~=~3.59$ for axis 1. This is because the combined approach optimally balances the disturbance parameters to ultimately converge to the smallest reachable tracking error.

\subsection{Our method vs. LQG control} \label{sec:eval:LQR}

The linear-quadratic-Gaussian control (LQG)  \cite{Doyle1981} is an optimization-based design approach, where the full state is estimated via a Kalman filter and state-feedback is generated, such that a cost function with weighting factors $Q$ and $r$ is minimized:
\begin{gather}\label{eq:eval:LQR}
	J = \sum_{k=0}^\infty (\vec{e}[k]^TQ\vec{e}[k] + u_p[k] r u_p[k]),
\end{gather}
where $\vec{e} = \vec{x}_r-\vec{x}_\mathrm{ref}$ is the state tracking error, and $\vec{x}_\mathrm{ref}[k] = [q_d[k-1],q_d[k],\dot{q}_d[k],\ddot{q}_d[k-1]]^T$ is the state reference. 
The Kalman filter assumes uncertainties in the model using zero-mean Gaussian noises with covariance matrices $S_W$ for the process and $S_V$ for the measurement, respectively. Here, we set $S_W = (1/3 \,\cdot G_{W_p} G_{W_p}^T)^2$ and $S_V = (1/3 \,\cdot G_{V_p} G_{V_p}^T)^2$, which assumes that the zero-centered sets $\mathcal{W}_p$ and $\mathcal{V}_p$ represent three times the standard deviation. We apply the \texttt{lqg} function from MATLAB and use our model from \eqref{eq:problem:robotLinearDynamics} for the design.

\begin{figure}
	\includegraphics[width=\columnwidth]{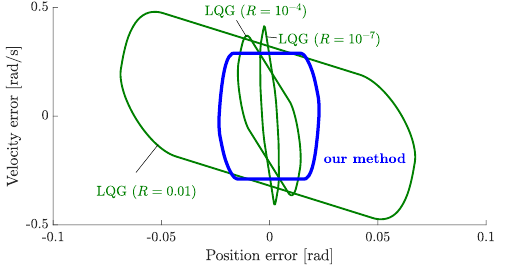}
	\caption{Comparison of the reachable tracking error $\mathcal{R}_e$ for controllers obtained with LQG optimization (green) and our method from Sec. \ref{sec:eval:separate} (blue).}
	\label{fig:eval:LQR}
\end{figure}

LQG relies on the user to set the weights in $Q$ and $r$. This is a difficult task, especially when there are input constraints to consider because, normally, the only way to determine whether a controller is feasible and desirable is to test it on the real system. Our paper realizes a different solution: using the reachset-conformant model from Tab. \ref{tab:eval:SF}, we can evaluate whether a possible weight combination may lead to an infeasible controller. To demonstrate this, we set $Q = \diag(1000,1000,0.01,r)$ and compute the reachable sets by varying $r$. 
For axis 1, we display the results in Tab. \ref{tab:eval:LQR} and the sets $\mathcal{R}_e$ are also visualized in Fig.~\ref{fig:eval:LQR}, including the reachable set obtained from Sec. \ref{sec:eval:separate} using combined synthesis.

\begin{table}
	\caption{Comparison: LQG optimization with $Q = \diag(1000,1000,0.01)$ vs. our~controller synthesis method for robot axis 1}
	\label{tab:eval:LQR}
	\setlength\tabcolsep{6pt}
	\renewcommand{\arraystretch}{1.25}
	\begin{center}
		\begin{tabular}{ r r r }
			\hline
			$R$ &$\|\mathcal{R}_e\|$ & $\mathcal{R}_u$ \\ \hline
        	$10^{-7}$ & $0.831$    & $[-75.37, 75.70]$\\  
        	\textbf{our method}&$\mathbf{0.625}$   & $\mathbf{[-23.24, 23.57]}$\\ 
       		$10^{-4}$& $0.762$   & $[-24.40, 24.72]$\\  
        	$0.01$&$1.091$  & $[-17.12, 17.45]$\\  
			\hline
		\end{tabular}
	\end{center}
\end{table}

The results show that if a high $r$ is set, then a weak controller is obtained, resulting in a large tracking error, but we receive the smallest $\mathcal{R}_u$ interval. The more $r$ is decreased, the more the tracking error improves. However, at $r$ near zero, the input constraints are violated. Instead, our optimization-based approach not only satisfies the input constraint but can use any controller and observer, while LQG is restricted to a state feedback design. As we described earlier, LQG requires test iterations to validate different combinations of possible $Q$ and $r$ and their resulting closed-loop performance, while our method requires test iterations only to make sure that the model remains conformant. As Sec. \ref{sec:eval:separate} showed, two iterations can be sufficient here.

\subsection{Comparing static feedback vs. disturbance-compensated feedback}\label{sec:eval:ESOvsHG}

\begin{figure}
	\includegraphics[width=\columnwidth]{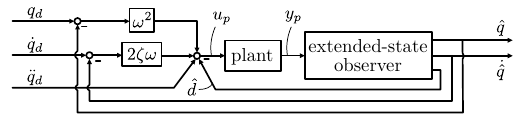}
	\caption{The control loop considered in Sec. \ref{sec:eval:ESOvsHG} tracks the reference $\vec{y}_\mathrm{ref}= [q_d,\dot{q}_d]$ and has an input feedforward $u_\mathrm{ff} = \ddot{q}_d$. An extended state observer \cite{Chen2016} is used to observe the position $q$, velocity $\dot{q}$, and the disturbance $d$, which are used in the feedback control.}
	\label{fig:eval:controllerSynthesis2}
\end{figure}

In the third experiment, we will demonstrate that our method is generalizable to other controllers besides the one specified in the previous two experiments. In the following, we synthesize an observer-based feedback control law with disturbance compensation
\begin{equation}\label{eq:experiments:feedbackLaw}
	u_p = \ddot{q}_d + \omega^2 (\dot{q}_d - \dot{\hat{q}}) + 2\zeta\omega (q_d-\hat{q}) - \hat{d},
\end{equation}
where $\hat{q},\dot{\hat{q}}$, and $\hat{d}$ are estimated by an extended-state observer (ESO) \cite{Chen2016}:
\begin{gather}
	\begin{bmatrix}\dot{\hat{q}}\\\ddot{\hat{q}}\\\dot{\hat{d}}\end{bmatrix} = \begin{bmatrix} 0 & 1 & 0 \\ 0 & 0 & 1 \\ 0 & 0 & 0 \end{bmatrix}\begin{bmatrix}\hat{q}\\\dot{\hat{q}}\\\hat{d}\end{bmatrix} +
	\begin{bmatrix}0\\1\\0\end{bmatrix} u_p + \begin{bmatrix}l_1/\epsilon\\l_2/\epsilon^2\\l_3/\epsilon^3\end{bmatrix}(q-\hat{q}).
\end{gather}

\begin{figure*}
	\includegraphics[width=178mm]{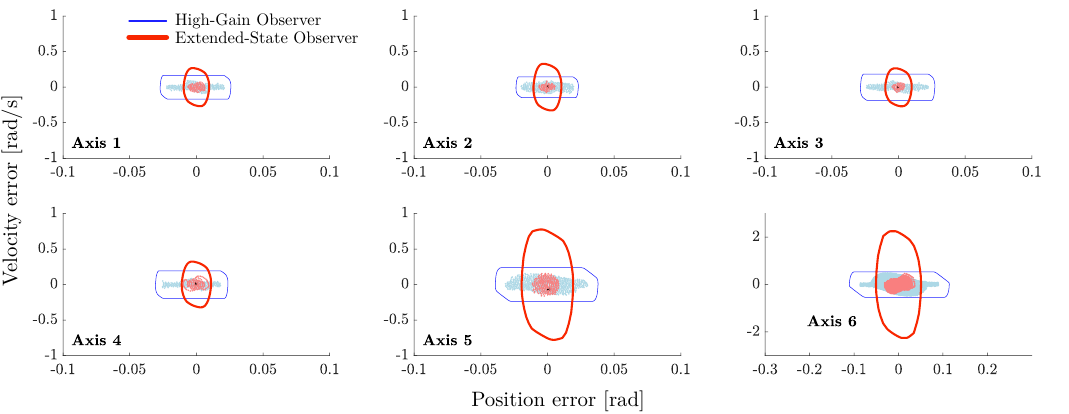}
	\caption{Comparison of the guaranteed tracking error $\mathcal{R}_e$ for controllers using a high-gain observer (blue) and an extended-state observer (red).}
	\label{fig:eval:ESOvsHG}
\end{figure*}

We compare this new controller against the one from the previous experiments. For the sake of brevity, we set $\omega=20$ and $\zeta = 1$ and only synthesize $h_1,h_2$ for the high-gain observer and $l_1,l_2,l_3$ for the extended-state observer, as well as the model uncertainties $\alpha_{W_p,1},\alpha_{W_p,2},\alpha_{V_p,1}$, respectively. We perform two iterations for each method: the results for the respective optimal parameters are shown in Tab. \ref{tab:eval:ESOvsHG} and the reachable set, as well as the measured tracking errors from the real robot, are shown in Fig. \ref{fig:eval:ESOvsHG}.

As the plotted reachable tracking errors show, the extended-state observers help to significantly improve the position error of the real robot, while the velocity error stays similar to the high-gain observer. As can be seen in the plots, the guarantees for the tracking error reflect a similar behavior. Axis 5 and 6 of our robot perform badly mainly due to insufficient feedback linearization. Nevertheless, the identified model remains conformant, and the reachable tracking error is correctly predicted. What is also noticeable is that the identified uncertain parameters $\alpha$ differ depending on the controller, e.g., for axis 6, $\alpha_{W_p,1}$ is larger for HG, while $\alpha_{W_p,2}$ is larger for ESO. One reason is that our controller synthesis chooses the optimal value that minimizes $\|\mathcal{R}_e\|$. Another reason is that the disturbance also depends on the controller since different controllers can suppress disturbances differently, e.g., the $\alpha_{W_p,2}$ are larger when using ESO, but the feedback law in \eqref{eq:experiments:feedbackLaw} is able to compensate for it, resulting in a smaller positional tracking error.

We summarize the experimental results of our combined controller synthesis and reachset-conformant identification. We demonstrated in Sec. \ref{sec:eval:separate} that a combined approach is necessary to avoid conservative results. In Sec. \ref{sec:eval:LQR}, we showed that LQG methods require careful balancing of the tracking error and the input effort, while our approach automatically satisfies the input constraints. In Sec. \ref{sec:eval:ESOvsHG}, we showed that our approach could be used for any controller structure as long as the closed-loop dynamics are linear. The experiment has also shown that although the observers do not consider the full dynamics of the plant, it is still possible to derive guarantees, and the soundness of our approach is not affected. Rather, we have shown for our robot that an observer with a better model may lead to a better performance of the closed-loop system.

\begin{table*}
	\caption{Comparing synthesized High-Gain observers with synthesized Extended-State Observers}
	\label{tab:eval:ESOvsHG}
	\setlength\tabcolsep{6pt}
	\renewcommand{\arraystretch}{1.25}
	\begin{center}
		\begin{tabular}{ r r r r r r r r r r r r r r r r }
			\hline
			& \multicolumn{6}{c}{\textbf{High-Gain Observer}}  && \multicolumn{7}{c}{\textbf{Extended-State Observer}} \\ 
			Axis& $\|\mathcal{R}_e\|$ & $h_1$ & $h_2$ & $\alpha_{W_p,1}$ & $\alpha_{W_p,2}$ & $\alpha_{V_p,1}$ && $\|\mathcal{R}_e\|$ & $l_1$ & $l_2$ & $l_3$ & $\alpha_{W_p,1}$ & $\alpha_{W_p,2}$ & $\alpha_{V_p,1}$ & \\\hline
			$1$&$0.373$&$135.0$&$416.1$&$0.0417$&$0.976$ &$8.73\cdot 10^{-6}$ &&$0.535$&$57.1$&$103.3$ &$18.1$ &$0.0234$&$3.587$ &$3.49\cdot 10^{-5}$\\
			$2$&$0.325$&$135.1$&$416.4$&$0.0397$&$0.652$ &$8.73\cdot 10^{-6}$ &&$0.662$&$93.5$&$182.0$ &$37.9$ &$0.0362$&$4.110$ &$3.49\cdot 10^{-5}$ \\
			$3$&$0.409$&$135.0$&$416.1$&$0.0539$&$0.679$ &$8.73\cdot 10^{-6}$ &&$0.534$&$75.6$&$113.7$ &$16.5$ &$0.0358$&$2.951$ &$3.49\cdot 10^{-5}$\\
			$4$&$0.428$&$135.0$&$416.1$&$0.0540$&$0.881$ &$8.73\cdot 10^{-6}$ &&$0.647$&$80.5$&$153.4$ &$30.1$ &$0.0283$&$4.415$ &$3.49\cdot 10^{-5}$\\
			$5$&$0.542$&$103.6$&$517.8$&$0.0344$&$2.780$ &$8.73\cdot 10^{-6}$ &&$1.573$&$37.6$&$490.0$ &$210.6$&$0.0055$&$17.936$&$8.73\cdot 10^{-6}$\\
			$6$&$1.297$&$105.5$&$602.4$&$0.0309$&$9.379$ &$8.73\cdot 10^{-6}$ &&$4.582$&$23.1$&$638.9$ &$338.7$&$0$     &$55.742$&$8.73\cdot 10^{-6}$  
		\\\hline
		\end{tabular}
	\end{center}
\end{table*}

%% file: sections/Discussion.tex
\section{Conclusion}\label{sec:conclusion}


In this paper, we have shown that our method can be used to optimally design a controller and to derive guarantees for the input constraint and tracking error. In contrast to previous work, these guarantees are also applicable to real robotic systems. Using our method, we can now formally analyze any linear robotic controller for their safety. 

The formal relation between the robot model and the real system is established by identifying reachset-conformant model parameters. The controller synthesis and identification are unified into a single optimization, which means that the model and controller are both optimized for the smallest reachable tracking error. 
Our experiments have shown that the computed reachable sets always successfully enclose all behaviors of the real robot system, however large the disturbance in the system is. Our approach does not require tuning of hyper-parameters, in contrast to LQR. We have shown the effectiveness of our novel approach to synthesizing different feedback laws. 

Our method can be applied to any robot in practice that uses feedback linearization, linear observers, and feedback controllers.
In the future, we would like to extend this approach to nonlinear plant models and controllers.

%% file: sections/Conclusion.tex


%% file: sections/Appendix.tex
{\appendices
{
\section{Full robot model including disturbance}\label{sec:appendix1}
To model the disturbance of the system, we make the following assumptions: 1) the velocity is disturbed by an interval $[-\alpha_{W_p,1},\alpha_{W_p,1}]$, 2) the acceleration is disturbed by an interval $[-\alpha_{W_p,2},\alpha_{W_p,2}]$, 3) the measurement is disturbed by an interval $[-\alpha_{V_p},\alpha_{V_p}]$, and 4) we consider an additional disturbance state, such that $\ddot{q} = u_p + d$ and $\dot{d} = 0$. The full model of the plant for each robot joint, including the uncertainties, is described by the following linear system:
\begin{align*}
	\vec{x}_p[k+1] &= \begin{bmatrix}
		0 & 1 & 0 & 0 & 0 \\ 
		0 & 1 & \Delta t & \Delta t^2/2 & \Delta t^2/2 \\
		0 & 0 & 1 & \Delta t & \Delta t \\
		0 & 0 & 0 & 1 & 0 \\
		0 & 0 & 0 & 0 & 0 
	\end{bmatrix} \vec{x}_p[k] + \begin{bmatrix}
		0 \\ 0 \\ 0 \\ 0 \\ 1
	\end{bmatrix} u_p[k]\\
	&+ \vec{w}_p[k], \\
	y_p[k] &= \begin{bmatrix}
		1 & 0 & 0 & 0 & 0
	\end{bmatrix} \vec{x}_p[k] + v_p[k],
\end{align*}
where $\vec{x}_p[k] = [q[k-1],q[k],\dot{q}[k],d[k],u_p[k-1]]^T$, $\vec{w}_p[k] \in \mathcal{W}_p$, and $\vec{v}_p[k] \in \mathcal{V}_{p}$:
\begin{equation*}
	\mathcal{W}_p = \left( \begin{bmatrix} 0 \\ 0 \\ 0 \\ 0 \\ 0\end{bmatrix}, 
	\begin{bmatrix}
		0 & 0\\ 
		\Delta t & \Delta t^2/2\\
		0 & \Delta t\\
		0 & 0\\
		0 & 0 
	\end{bmatrix}
	\begin{bmatrix}
		\alpha_{W_p,1} \\ 
		\alpha_{W_p,2}
	\end{bmatrix}
	\right),
	\mathcal{V}_p = \left(0,
	\alpha_{V_p}
	\right),
\end{equation*}
where $\alpha_{W_p,1}$, $\alpha_{W_p,2}$, and $\alpha_{V_p}$ are the scaling factors of the zonotopes $\mathcal{W}_{p}$ and $\mathcal{V}_{p}$. The generator matrix of $\mathcal{W}_{p}$ is a discretization similar to the plant linear dynamics.
}
{
\section{Analysis of the reference error}\label{sec:appendix2}
The vector $\tilde{\vec{y}}_\mathrm{cl,ref}$ is computed using \eqref{eq:problem:reachableSet} considering $\vec{y}_\mathrm{ref}$ and $u_\mathrm{ff}$ and considering $\vec{w}_p = \vec{0}, v_p = 0$. The result is a trajectory that tracks the reference with a \textit{reference error}, which we define as $\vec{y}_\mathrm{e,ref}$ and $u_\mathrm{e,ff}$ such that
\begin{equation}
	\tilde{\vec{y}}_\mathrm{cl,ref} = \begin{bmatrix}
		\vec{y}_\mathrm{ref} + \vec{y}_\mathrm{e,ref} \\
		u_\mathrm{ff} + u_\mathrm{e,ff}
	\end{bmatrix}.
\end{equation}
In cases where $u_\mathrm{ff}$ is the output of the inverted plant model \cite{Moylan1977} given $y_\mathrm{ref}$ as an input, there will be no reference error. A simple example is a double-integrator model $\ddot{q} = u_p$, where the output is $\vec{y}_p=[q,\dot{q}]^T$. Applying $u_p=\ddot{q}_d$ would exactly produce the reference $\vec{y}_p = [q_d,\dot{q}_d]$ without any error. In any other case, the tracking error increases by $\vec{y}_\mathrm{e,ref}$ and thus requires an additional input $u_\mathrm{e,ff} = -K\vec{y}_\mathrm{e,ref}$, where $K = [\omega^2,2\zeta\omega]$, to compensate for the additional tracking error. In some cases, the additional input could lead to a violation of the input constraint: $u_\mathrm{e,ff} \oplus \mathcal{R}_u \not\subset \mathcal{U}_p$. In the following paragraphs, we present three different ways to deal with the reference error to arrive at an actual reachable tracking error $\mathcal{R}^*_e$ and reachable input $\mathcal{R}^*_u$:
\subsubsection{Tracking $\vec{y}_\mathrm{cl,ref}$ instead of $\vec{y}_\mathrm{ref}$}
Let us rewrite the control law in \eqref{eq:eval:staticfeedback}, considering $\vec{y}_\mathrm{ref} = [q_d,\dot{q}_d]^T$, $\vec{y}_\mathrm{cl} = [\hat{q},\dot{\hat{q}}]^T$, $K = [\omega^2, 2\zeta\omega],$ and $\vec{y}_e \in \mathcal{R}_e$ such that
\begin{align*}
	u_p &= u_\mathrm{ff} + K(\vec{y}_\mathrm{ref} - \vec{y}_\mathrm{cl}) \\
	&= u_\mathrm{ff} + K(\vec{y}_\mathrm{ref} - (\vec{y}_\mathrm{ref} + \vec{y}_\mathrm{e,ref}  +  \vec{y}_e))\\
	&= u_\mathrm{ff} - K(\vec{y}_\mathrm{e,ref}  +  \vec{y}_e), = u_\mathrm{ff} + u_\mathrm{e,ff}  -K\vec{y}_e.
\end{align*}
We slightly modify the static-feedback control law to track $\vec{y}_\mathrm{cl,ref}$ instead of $\vec{y}_\mathrm{ref}$ such that
\begin{align*}
	u_p^* &= u_\mathrm{ff} + K(\vec{y}_\mathrm{cl,ref} - \vec{y}_\mathrm{cl}) \\
	&= u_\mathrm{ff} + K(\vec{y}_\mathrm{ref} + \vec{y}_\mathrm{e,ref} - (\vec{y}_\mathrm{ref} + \vec{y}_\mathrm{e,ref}  +  \vec{y}_e))\\
		&= u_\mathrm{ff}   -K\vec{y}_e.
\end{align*}
By definition, $u_\mathrm{e,ff}$ vanishes using the new control law, and the input constraint cannot be violated anymore. Since no uncertainty is involved in obtaining $\vec{y}_\mathrm{cl,ref}$ and $\vec{y}_\mathrm{e,ref}$, they can be exactly precomputed before executing a trajectory. The actual reachable sets are then defined as
\begin{align*}
	\mathcal{R}_e^*&:=\vec{y}_\mathrm{e,ref} \oplus \mathcal{R}_e, \\
	\mathcal{R}_u^*&:=\mathcal{R}_u.
\end{align*}
The advantage is that the input constraint is guaranteed independently of the desired trajectory. The disadvantage, however, is that we deviate from the original control law, and that $\mathcal{R}_e$ is relative to $\vec{y}_\mathrm{cl,ref}$ instead of $\vec{y}_\mathrm{ref}$.
\subsubsection{Precomputing the reference error}
As no uncertainty is involved, $\vec{y}_\mathrm{e,ref}$ and $u_\mathrm{e,ff}$ can be precomputed before executing a trajectory. We define the actual reference-dependent sets as
\begin{align*}
	\mathcal{R}_e^*&:=\vec{y}_\mathrm{e,ref} \oplus \mathcal{R}_e, \\
	\mathcal{R}_u^*&:=u_\mathrm{e,ff} \oplus \mathcal{R}_u.
\end{align*}
The disadvantage of this approach is, however, that the input constraint cannot be guaranteed at all times; $\mathcal{R}_u^* \subseteq \mathcal{U}_p$ must be checked before every execution of a trajectory on the robot. We only recommend this approach if the controller is designed for a single reference trajectory.
\subsubsection{Solve \eqref{eq:problem:optimProblem} for a predefined set of references}
In this approach, we predefine a large set of reference trajectories before solving \eqref{eq:problem:optimProblem}, e.g., we can use the same trajectories from the test cases used to identify the disturbances. We then compute two sets $\mathcal{Y}_\mathrm{e,ref}$ and $\mathcal{U}_\mathrm{e,ff}$, that enclose all $\vec{y}_\mathrm{e,ref}$ and $u_\mathrm{e,ff}$ for all references.
 The actual reachable sets are then defined as
 \begin{align*}
	\mathcal{R}^*_e &:= \mathcal{Y}_\mathrm{e,ref} \oplus \mathcal{R}_e, \\
	\mathcal{R}^*_u &:= \mathcal{U}_\mathrm{e,ff} \oplus \mathcal{R}_u,
\end{align*}
and replace $\mathcal{R}_e$ and $\mathcal{R}_u$ when solving \eqref{eq:problem:optimProblem}. The advantage is that the input constraint is guaranteed for all considered references, and also all non-considered references where $u_\mathrm{ff} \in \mathcal{U}_\mathrm{e,ff}$, while the effort for solving \eqref{eq:problem:optimProblem} is only slightly increased. We used this method in our experiments in Sec. \ref{sec_evaluation}. We recommend this approach if the controller is designed for unknown references, but when the method for reference generation stays similar, e.g., always $u_\mathrm{ff}[k] = \ddot{q}_d[k]$, or $u_\mathrm{ff}[k] = \ddot{q}_d[k+2]$ to consider delays in the plant. However, during pre-computation, sufficiently many reference trajectories are necessary so that the largest possible $\mathcal{U}_\mathrm{e,ff}$ can be found. 
}